\documentclass{article}

\usepackage{microtype}
\usepackage{graphicx}
\usepackage{booktabs} 

\usepackage{hyperref}

\usepackage{wrapfig}
\usepackage{subfig}

\usepackage[accepted]{icml2024}

\usepackage{amsmath}
\usepackage{amssymb}
\usepackage{mathtools}
\usepackage{amsthm}

\usepackage[capitalize,noabbrev]{cleveref}

\theoremstyle{plain}
\newtheorem{theorem}{Theorem}[section]

\newtheorem{corollary}[theorem]{Corollary}
\theoremstyle{definition}
\newtheorem{definition}[theorem]{Definition}

\theoremstyle{remark}
\newtheorem{remark}[theorem]{Remark}
\newtheorem*{theorem*}{Theorem}
\newtheorem*{corollary*}{Corollary}
\theoremstyle{definition}

\usepackage[textsize=tiny]{todonotes}

\icmltitlerunning{On the Recoverability of Causal Relations from Temporally Aggregated I.I.D. Data}

\begin{document}

\twocolumn[
\icmltitle{On the Recoverability of Causal Relations from \\Temporally Aggregated I.I.D. Data}



\icmlsetsymbol{equal}{*}

\begin{icmlauthorlist}
\icmlauthor{Shunxing Fan}{mbzuai}
\icmlauthor{Mingming Gong}{um,mbzuai}
\icmlauthor{Kun Zhang}{mbzuai,cmu}
\end{icmlauthorlist}

\icmlaffiliation{mbzuai}{Department of Machine Learning, Mohamed bin Zayed University of Artificial Intelligence, Abu Dhabi, United Arab Emirates}
\icmlaffiliation{cmu}{Department of Philosophy, Carnegie Mellon University, Pittsburgh, PA, United States}
\icmlaffiliation{um}{School of Mathematics and Statistics, The University of Melbourne, Melbourne, Australia}

\icmlcorrespondingauthor{Kun Zhang}{kunz1@cmu.edu}

\icmlkeywords{Causality, Causal Discovery, Machine Learning, ICML}

\vskip 0.3in
]



\printAffiliationsAndNotice{} 

\begin{abstract}

We consider the effect of temporal aggregation on instantaneous (non-temporal) causal discovery in general setting. 
This is motivated by the observation that the true causal time lag is often considerably shorter than the observational interval. This discrepancy  leads to high aggregation, causing time-delay causality to vanish and instantaneous dependence to manifest. Although we expect such instantaneous dependence has consistency with the true causal relation in certain sense to make the discovery results meaningful, it remains unclear what type of consistency we need and when will such consistency be satisfied. We proposed functional consistency and conditional independence consistency in formal way correspond functional causal model-based methods and conditional independence-based methods respectively and provide the conditions under which these consistencies will hold. We show theoretically and experimentally that causal discovery results may be seriously distorted by aggregation especially in complete nonlinear case and we also find causal relationship still recoverable from aggregated data if we have partial linearity or appropriate prior.
Our findings suggest community should take a cautious and meticulous approach when interpreting causal discovery results from such data and show why and when aggregation will distort the performance of causal discovery methods.

\end{abstract}

\section{Introduction}

Causal discovery methods, which aim to uncover causal relationships from observational data, have been extensively researched and utilized across multiple disciplines including computer science, economics, and social sciences~\citep{pearl2009causality,spirtes2000causation}. 
These methods can be broadly categorized into two types based on whether they explicitly utilize temporal information of the data. The first type is temporal causal discovery, which is specifically designed for analyzing time series data. Examples of this type include the Granger causality test \citep{granger1969investigating} and its variants. The second type is non-temporal causal discovery referring to methods that do not explicitly utilize time index information, which is applicable to independent and identically distributed (i.i.d.) data. This category encompasses various approaches such as constraint-based, score-based, and functional causal model (FCM)-based methods like PC \citep{spirtes2000causation}, GES \citep{chickering2002optimal}, and LiNGAM \citep{shimizu2006linear}.

Temporal causal discovery often relies on the assumption that the causal frequency aligns with the observation frequency, but in real-world scenarios the causal frequency is often unknown, which means that the available observations may have a lower resolution than the underlying causal process. An instance of this is annual income, which is an aggregate of monthly or quarterly incomes \citep{drost1993temporal}. Moreover, it is widely believed that causal interactions occur at  high frequencies in fields such as economics \citep{ghysels2016testing} and neuroscience \citep{zhou2014analysis}. Some research has been conducted to explore the effects of temporal aggregation on time series modeling \citep{ghysels2016testing,marcellino1999some,silvestrini2008temporal,granger1999effect,rajaguru2008temporal,gong2017causal}. These works typically consider small aggregation factor $k$ and still treat the temporal aggregation from causal processes as a time series \footnote{The "aggregation factor $k$" refers to the number of data points from the underlying causal process that are combined to form each observed data point. It is also called the aggregation level or aggregation period.}.

However, in many real-world scenarios, the temporal aggregation factor can be quite large (we only consider aggregation without overlapping windows). In these cases, what was originally a time-delayed dependence can appear as an instantaneous dependence when observed. An example commonly used in statistics and econometrics textbooks to illustrate correlation, causation, and regression analysis is the influence of temperature on ice cream sales.
Intuitively, one might think that the average daily temperature has an instantaneous causal effect on the total daily ice cream sales. However, in reality, the causal process involves a time lag: a high temperature at a specific past moment influences people's decision to purchase ice cream, which then leads to a sales transaction at a subsequent moment. But we typically work with the aggregation of this causal interaction, such as the average daily temperature and the total daily ice cream sales, which represent the sum of all individual sales transactions over the day and that is the reason we observed instantaneous causality.

Interestingly, the causality community  has long acknowledged the significance of temporal aggregation as a common real-world explanation for instantaneous causal models like the structural equation model.
\citet{fisher1970correspondence} argued that simultaneous equation models serve as approximations of true time-delayed causal relationships driven by temporal aggregation in the limit. He emphasized that while causation inherently involves a temporal aspect, as the reaction interval tends to zero, the aggregation factor $k$ tends to infinity. 
\citet{granger1988some} shared a similar view and claimed that ``temporal aggregation is a realistic, plausible, and well-known reason for observing apparent instantaneous causation''. This explanation has been consistently used in recent causal discovery papers, especially those discussing cyclic models~\citep{rubenstein2017causal,lacerda2012discovering,hyttinen2012learning}. 

When applying non-temporal causal discovery methods to uncover instantaneous causal relationships resulting from temporal aggregation, a fundamental question arises: Are these estimated instantaneous causal relationships consistent with the true time-delayed causal relationships? 
This issue is crucial because our primary concern lies in discerning the true causal relations. If the results obtained by the instantaneous causal discovery methods do not align with the true causal relationship, the results may not suggest anything like how to apply intervention. Regrettably, few studies have examined the alignment of these estimated instantaneous causal relationships stemming from temporal aggregation with the true time-delayed causal relationships. To the best of our knowledge, the only theoretical analysis  related to this question is conducted by \citet{fisher1970correspondence} and \citet{gong2017causal}. We will delve into a comprehensive discussion of their contributions in subsection \ref{related work}.

\subsection{Contributions}
We propose and investigate two different sense of principle consistencies: functional consistency and conditional independence consistency to formally define what does the recoverability means when we perform FCM-based and conditional independence-based methods on aggregated data respectively. To the best of our knowledge, our paper is the first to specifically discuss the risks and feasibility of performing non-temporal causal discovery methods on aggregated data in general(nonlinear) cases.
\begin{enumerate}
    \item Functional consistency requires that the aggregated model maintain the consistent functional form of the original causal model. We discuss functional consistency in the context of additive noise models and non-stationary scenarios. For the additive noise model, we present the construction form of the causal function after aggregation in Theorem \ref{thm:ConstructionOfHatF} and also provide the necessary and sufficient conditions for the Structural Causal Model (SCM) to hold. For non-stationary scenarios, we discuss different cases and provide results for the general case in Theorem \ref{thm:GeneralCaseForFunctionalConsistency}.
    \item Conditional independence consistency expects aggregated model maintain the consistent conditional independence properties as the original causal model. We investigate the three fundamental trivariate structure, chain, fork and collider. We show the collider structure naturally has conditional independence consistency with the causal Markov condition, even for the nonlinear case, but chain and fork do not (see Figure \ref{fig:aligned_models} and Remark \ref{remark: chainlike not consistent in general}). And we provide the necessary and sufficient condition(Theorem \ref{ci main theorem}) for the conditional independence consistency for chain and fork structure which can be presented as a equation involving conditional density functions. Each conditional density functions only involving two components of the time series, which motivate us to prove that partial linearity is sufficient to ensure such consistency for chain and fork(Corollary \ref{Sufficient Conditions for Conditional Independence} and \ref{corollary:partial linear is enough}).
    \item We conducted five simulation experiments to support our findings, focusing on theorems for functional and conditional consistency, examining the effect of the $k$ value, and proposing a trivial solution for the aggregation issue.
\end{enumerate}

\subsection{Related work}\label{related work}

\citet{fisher1970correspondence} established the corresponding relationship between the simultaneous equation model and the true time-lagged model, providing the conditions to ensure such correspondence. His analysis encompassed both linear and general cases. Roughly speaking, he conducted theoretical analysis to show that this correspondence can be ensured when the function of the equation has a fixed point. However, the assumptions he employed were quite restrictive, assuming that the value of noise is fixed for all the causal reactions during the observed interval. Some subsequent studies have also adopted this assumption \citep{rubenstein2017causal,lacerda2012discovering}. This assumption is too strong, as it implies that noise in the causal reaction is only related to our observation like measurement error. Actually, the noise defined in structural causal models or functional causal models also represents unobserved or unmeasured factors that contribute to the variation in a variable of interest, rather than being merely observational noise.

\citet{gong2017causal} adopted a more practical and flexible assumption in their work. They defined the original causal process as a vector auto-regressive model (VAR) $X_{t}=AX_{t-1}+e_t$ and allowed for the randomness of the noise $e_t$ in the observation interval. They gave a theoretical analysis showing that, in the linear case and as the aggregation factor tends to infinity, the temporally aggregated data $\overline{X}$ becomes i.i.d. and compatible with a structural causal model $\overline{X}=A\overline{X}+\overline{e}$. In this model, matrix $A$ is consistent with the matrix $A$ in the original VAR. This suggests that high levels of temporal aggregation preserve functional consistency in the linear case. However, their study only considers the linear case and lacks analysis for general cases.

\section{Preliminary}

In this section, we first introduce the Vector Autoregressive (VAR) model and its temporal aggregation. Then, we briefly describe how, as the aggregation level $k$ increases, the time-delay dependence vanishes and how it can be approximated as the aggregation of an instantaneous causal model. Based on this fact, we ignore the unaligned variables caused by time-delay causal effects and define the underlying process with instantaneous causal effects across different components, while allowing potential time-delay causal effects between variables within the same component. The analysis in subsequent sections is based on the aggregation of instantaneous causal effects.

\subsection{VAR Model and its Temporal Aggregation}
Aligning with the settings in \citet{gong2017causal}, we assume the underlying causal process can be described by a VAR(1) (autoregressive model with a maximum lag of 1):
\begin{equation*}
    X_t = f(X_{t-1}, N_t),\quad t\geq2,
\end{equation*}
where $X_t = (X_t^{(1)}, X_t^{(2)}, \dots, X_t^{(s)})^T$ is the observed data vector at time $t$, $s$ is the dimension of the random vector. $f$ is a vector-valued function $\mathbb{R}^{2s} \to \mathbb{R}^s$, and $N_t=(N_t^{(1)}, \dots, N_t^{(s)})^T$ denotes a temporally and contemporaneously independent noise process. When mentioning VAR in our paper, we refer to the general VAR model defined above, which includes both linear and nonlinear functions.

The temporally aggregated time series of this process, denoted by $\overline{X_t}$, is defined as: 
\begin{equation}
    \overline{X_t}=\frac{\sum_{i=1}^{k} X_{i+(t-1)k}}{g(k)}.
    \label{eq:aggregrate}
\end{equation} 

When we mention the aggregation of the time-delay model in the paper, we consider cases where $k$ is large by default. In such cases, we treat the temporally aggregated data as i.i.d. data, and we will drop the subscripts $t$ in Eq. \ref{eq:aggregrate} from now on. The normalization factor $g(k)$ is generally required to satisfy $\lim\limits_{k \to \infty} g(k) = +\infty$, like $g(k)=k$. It should be chosen carefully to ensure that the limit of temporal aggregation has a finite, non-zero variance.

\begin{figure}[ht]
\centering
\begin{minipage}{0.8\linewidth} 
  \centering
  \includegraphics[width=\linewidth]{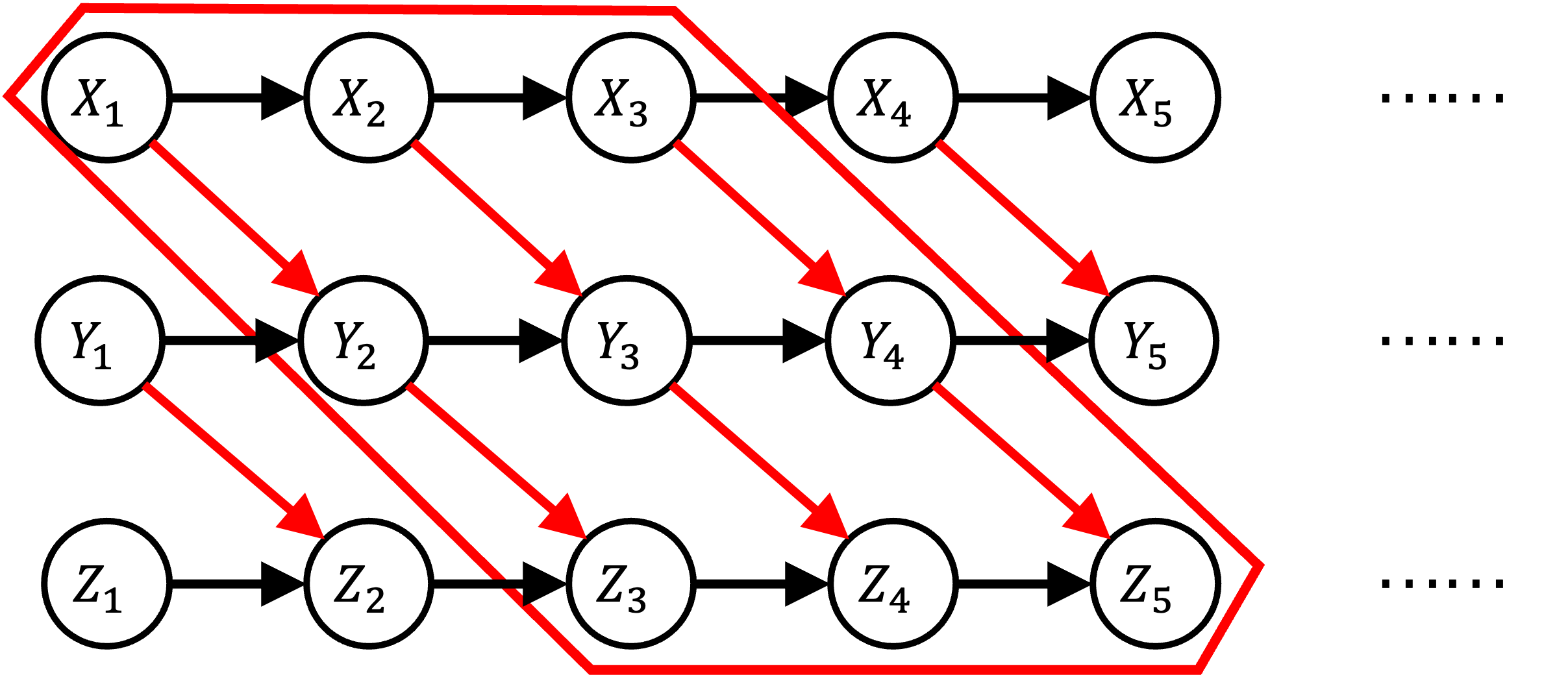}
\end{minipage}%
\hspace{0.05\linewidth} 
\begin{minipage}{0.08\linewidth} 
  \centering
  \includegraphics[width=\linewidth]{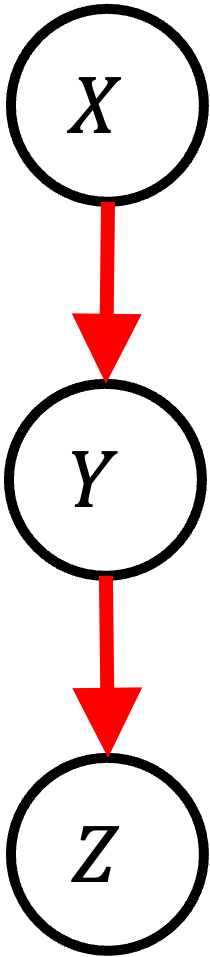}
\end{minipage}

\caption{Left: Directed acyclic graph for the VAR model with chain-like cross lag effects. Right: The corresponding summary graph.}
\label{fig:combined_graph} 
\end{figure}

We also introduce the concept of a summary graph, which often serves as the objective for causal discovery from a random process. Figure \ref{fig:combined_graph} shows an example of a summary graph.

\begin{definition}[Summary Graph]
Each time series component is collapsed into a node to form the summary causal graph. The summary graph represents causal relations between time series without referring to time lags. If there is an arrow from $X_{t-k}^{(i)}$ to $X_{t}^{(j)}$ in the original process for some $k\geq 0$, then it is represented in the summary graph.
\end{definition}

\subsection{Approximation}

In this subsection, we introduce \textbf{how the aggregation of a time-delay model can be approximated as the aggregation of an instantaneous model.} This section merely provides a simple example to illustrate why the aggregation of instantaneous causation can be used to approximate the aggregation of time-delay causation. In the Appendix \ref{Theorem and Proof for Approximating}, we provide more rigorous and general theorem and proof to justify this approach.

See Figure \ref{fig:combined_graph} as an example. The summary graph of this VAR is a chain structure involving triple variables. But the chain structure actually occurs in a lagged form: $X_t \rightarrow Y_{t+1} \rightarrow Z_{t+2}$ \footnote{Here $X_t$ represents a one-dimensional variable. Starting from here, $X_t$ and $\overline{X}$ represent scalar random variable, and we will use X, Y, Z,... to represent the different components of multivariate time series, instead of using $X^{(1)}$,...,$X^{(s)}$ as defined in \ref{section: functional subsection: def}. }. For analytical convenience, we will perform an alignment $X'_t:=X_t,Y'_{t}=Y_{t+1},Z'_{t}:=Z_{t+2}$ to make the causal effect instantaneous. We refer to the instantaneous model over $X'_t,Y'_{t},Z'_t:$ as the aligned model. In the example of chain-like VAR, the aligned model is in the red box in the Figure \ref{fig:combined_graph}.

When $k$ is large, the temporal aggregation from the original VAR and the aligned model is exactly the same: as $k$ approaches infinity, the temporally aggregated data $(\overline{X'}, \overline{Y'}, \overline{Z'})$ tends towards $(\overline{X}, \overline{Y}, \overline{Z})$. This can be demonstrated by the following equalities: Since $g(k)\rightarrow \infty$, we have
\begin{align*}
\overline{Y'} - \overline{Y} &= \frac{\sum_{i=2}^{k+1} Y_i}{g(k)} - \frac{\sum_{i=1}^k Y_i}{g(k)} = \frac{Y_{k+1} - Y_1}{g(k)} \rightarrow 0, \\
\overline{Z'} - \overline{Z} &= \frac{\sum_{i=3}^{k+2} Z_i}{g(k)} - \frac{\sum_{i=1}^k Z_i}{g(k)} \\&= \frac{Z_{k+2} + Z_{k+1} - Z_1 - Z_2}{g(k)} \rightarrow 0,
\end{align*}
as $k \rightarrow \infty$.

This approximation is natural, as the similar assertion that high aggregation causes time-delay dependence to become instantaneous has appeared and been discussed in many papers~\citep{pierce1977causality,granger1980testing,granger1988some,renault1998testing,breitung2002temporal,gong2017causal}. Please note that the number of unaligned variables we rule out (outside the red block in Figure \ref{fig:combined_graph}) is quite small compared to the variables within the block. Therefore, as $k$ increases, the aggregation of the time-delay model will rapidly approach the aggregation of the instantaneous model in practice (see Appendix \ref{Effect of k Value appendix} for an experiment investigating the effect of $k$ value). If the $k$ value is quite small and the unaligned variables significantly contribute to causal discovery, it indicates that the data exhibit an obvious time-delay property. In such cases, one should employ a temporal causal discovery method, which is beyond the scope of this paper.

\textbf{Therefore, all the theoretical results in this paper consider the aggregation of the instantaneous underlying model, also referred to as the aligned model.} They are formally defined in Definition \ref{def:Bivariate Aligned Model with Instant Structures} in Section \ref{section:Functional Consistency} and Definition \ref{def: Aligned Model with Instant Structures} in Section \ref{section:Conditional Independence Consistency}. From this perspective, our paper can be seen as exploring when it is possible to recover causal relations from the aggregation of instantaneous causal models and \textbf{all the results can be applied to them with any finite $k$ value and any choice of $g(k)$} (thus we set $g(k)=1$ for simplicity in the subsequent sections). Only when we want to apply the results to the aggregation of the time-delay model do we need $k$ to be large and $g(k)$ to be appropriate, i.e., $\lim\limits_{k \to \infty} g(k) = +\infty$.

\section{Functional Consistency: Recoverability with FCM-based Methods}\label{section:Functional Consistency}
FCM-based methods make stronger assumptions and utilize more information beyond just conditional independence. Thus, they can distinguish cause from effect from observational data under the functional assumptions. If we want to ensure the reliability of the results from FCM-based causal discovery on temporally aggregated data, we need some functional consistency between the process of temporally aggregated data and the true causal process.


\subsection{Definitions}\label{section: functional subsection: def}

The FCM-based methods primarily focus on the causal direction and consider the bivariate case in their studies. Aligned with~\citep{shimizu2011directlingam,hoyer2008nonlinear,zhang2012identifiability}, we also consider the process involving two components. 

\begin{definition}[Bivariate Aligned Model with Instant Structures]\label{def:Bivariate Aligned Model with Instant Structures}
The aligned model for VAR model with two components $X\to Y$ is defined below:

$X_0$, $Y_0$ are independent and follow the initial distribution.
when $t\geq1$,

$X_{t}=f_X(X_{t-1},N_{X,t})$, $Y_{t} = f_Y(X_t, N_{Y,t})$

where $f_X, f_Y$ are general functions. $N_{X,t}$, $N_{Y,t}$ are independent random variables with non-zero variance, which are independent of each other and they are identically distributed and independent across time t.

The temporal aggregation (for simplicity we set $g(k)=1$) are denoted as $\overline{X} := \sum_{i=1}^k X_i$, and similarly for $\overline{Y}$, $\overline{Z}$. 

\end{definition}

Intuitively, functional consistency implies that the functional causal model still exists after aggregation and adheres to the correct causal direction. We attempt to formulate a general definition; however, the problem becomes trivial because a causal function and independent noise can always be constructed to create a Structural Causal Model between any two random variables if no additional constraints are applied~\citep{darmois1953analyse,hyvarinen1999nonlinear,zhang2015estimation}. Therefore, in line with the requirements of commonly used FCM-based methods, we introduce some reasonable constraints, such as requiring the model to incorporate additive noise.

\subsection{Discussion for Two Types of Functional Consistency}
\begin{definition}[Functional Consistency Regarding Additive Noise]\label{def:FunctionalConsistencyRegardingAdditiveNoise}
Consider the bivariate aligned model defined in \ref{def:Bivariate Aligned Model with Instant Structures} incorporates additive noise: \(Y_t = f(X_t) + N_{Y,t}\). This process exhibits functional consistency regarding additive noise if there exists a function \(\hat{f}\) such that the aggregated variables can be represented as \(\overline{Y} = \hat{f}(\overline{X}) + N\), where \(N\) is independent of \(\overline{X}\), and such \(\hat{f}\) exists only in the correct causal direction.
\end{definition}

When the original causal function \(f\) is linear, then \(\overline{Y} = \sum_{i=1}^{k} (AX_i + N_{Y,t}) = A\overline{X} + \sum_{i=1}^{k} N_{Y,i}\) ~\citep{gong2017causal}, \(\sum_{i=1}^{k} N_{Y,i}\) is independent of \(\overline{X}\). However, because such an SCM exists on both sides, it is unidentifiable; we require non-Gaussian~\citep{shimizu2006linear,shimizu2011directlingam} or nonlinearity~\citep{hoyer2008nonlinear} to achieve asymmetry. Nonetheless, due to the central limit theorem, even if \(N_{Y,i}\) is non-Gaussian, \(\sum_{i=1}^{k} N_{Y,i}\) will quickly exhibit Gaussian characteristics, making it unidentifiable. See Appendix \ref{functional consistency appendix} for experiments showing how rapidly non-Gaussianity vanishes as \(k\) increases.

In the nonlinear case, we derive the following theorem:

\begin{theorem}[Construction of $\hat{f}$]\label{thm:ConstructionOfHatF}
If such $\hat{f}$, as defined in Definition \ref{def:FunctionalConsistencyRegardingAdditiveNoise}, exists, then $\hat{f}$ must take the form:
\begin{equation}\label{eq:ConstructionOfHatF}
    \hat{f}(T) = \mathbb{E}\left(\sum_{i=1}^k f(X_i) \mid \overline{X}=T\right) + c,
\end{equation}
where $c$ is any constant (which can be incorporated into the noise term) and the expression \(\mathbb{E}(\cdot \mid \overline{X}=T)\) denotes the conditional expectation. For simplicity, we set $c=0$. Consequently, this implies:
\begin{equation}
    \mathbb{E}(\hat{f}(\overline{X})) = \mathbb{E}\left(\sum_{i=1}^k f(X_i)\right).
\end{equation}
\end{theorem}

See Appendix \ref{proof:additive noise constrction} for the proof. This construction of $\hat{f}$ is very interesting and bears a strong resemblance to the Rao-Blackwell Theorem ~\citep{lehmann2006theory}, which aims to find an estimator (or a function in the causal inference framework) that accurately represents the underlying relationship between variables while accounting for noise or other confounding factors. The principle that the new estimator should be a function of some statistics, and using conditional expectations to enforce this, is very similar to what is described by the Theorem \ref{thm:ConstructionOfHatF}.

Then we can present $\overline{Y}$ in two ways: $\overline{Y} = \sum_{i=1}^k f(X_i) + \sum_{i=1}^k N_{Y,i} = \hat{f}(\overline{X}) + N$, where $\hat{f}$ is known and, according to the definition, we expect $N$ to be independent of $\overline{X}$. Therefore, we directly imply the results below.

\begin{theorem}[Necessary and Sufficient Condition]\label{thm:NecessaryAndSufficientCondition}
The necessary and sufficient condition for the existence of the additive noise causal model defined in Definition \ref{def:FunctionalConsistencyRegardingAdditiveNoise} is that $N = \sum_{i=1}^k N_{Y,i} + \left( \sum_{i=1}^k f(X_i) - \hat{f}(\overline{X}) \right)$ is independent of $\overline{X}$,
where $\hat{f}$ is defined by Eq. \ref{eq:ConstructionOfHatF}.
\end{theorem}

The proof can be found in Appendix \ref{proof for necessary and sufficient condition for additive}.

It is worth noting that the construction of $\hat{f}$ naturally ensures that $N$ is uncorrelated with $\overline{X}$. However, it is still challenging to achieve independence. For the condition, we find that the term $\sum_{i=1}^k N_{Y,i}$ is independent of every $X_i$ and $\overline{X}$, as well as the construction of $\hat{f}$. Thus, we cannot rely on this term to enforce independence. Furthermore, the term $\sum_{i=1}^k f(X_i) - \hat{f}(\overline{X})$ is likely dependent on $\overline{X}$. If we calculate the conditional variance, conditional on $\overline{X} = T$, for $\sum_{i=1}^k f(X_i) + \sum_{i=1}^k N_{Y,i}$ and $\hat{f}(\overline{X}) + N$ respectively, then we will get $\text{var}(N) = \text{var}\left[\sum_{i=1}^k N_{Y,i}\right] + \text{var}\left[\sum_{i=1}^k f(X_i) \mid \overline{X} = T\right]$. Please note that the LHS is not related to $T$, and the first term in RHS is not related to $T$ as well. This requires $\text{var}\left[\sum_{i=1}^k f(X_i) \mid \overline{X} = T\right]$ also to be unrelated to $T$, which is a non-trivial requirement for the distribution of $X_i$ and $f$.

Sometimes, data is collected from different regions/modules. The causal mechanism remains unchanged across these regions, but the distribution of data may vary. The identifiability of some causal discovery methods depends on the variance in distribution and the consistency of the causal function across different regions~\citep{huang2020causal}. We are now investigating whether the same causal mechanism persists in different regions when considering aggregated data.

\begin{definition}[Functional Consistency with respect to different regions]\label{def:Functional Consistency with respect to different regions}
Consider causal models $X_{t,A}=f_X(X_{t-1,A},N_{X,t,A})$, $Y_{t,A} = f(X_{t,A}, N_{t,A})$ and $X_{t,B}=f_X(X_{t-1,B},N_{X,t,B})$, $Y_{t,B} = f(X_{t,B}, N_{t,B})$ for regions A and B, respectively, where $N_{X,t,A}$, $N_{X,t,B}$, $N_{t,A}$, and $N_{t,B}$ represent independent noises with region-specific distributions and are i.i.d. within each region. \(\overline{X}_{A/B}\) and \(\overline{Y}_{A/B}\) are the aggregated data across individuals within each region. This process exhibits functional consistency with respect to different regions if, only for the correct causal direction  \(X \to Y\), there exists a function \(\hat{f}\) such that the aggregated variables satisfy \(\overline{Y}_A = \hat{f}(\overline{X}_A, N_A)\) and \(\overline{Y}_B = \hat{f}(\overline{X}_B, N_B)\).
\end{definition}

Obviously, if $f$ is a linear function, then it will always be consistent across different regions. For the nonlinear additive noise model, according to the construction formula Eq. \ref{eq:ConstructionOfHatF}, which is rely on the distribution of $X_i$, it is feasible to provide examples of $\hat{f}$ being inconsistent across different regions (see Appendix \ref{Example for inconsistency}). In general cases, due to a lack of constraint, it is always possible to find a consistent function; however, such a function can exist in both directions, rendering the system still unidentifiable. See Appendix \ref{proof for diffrent regions} for the proof.

\begin{theorem}[General Case for Functional Consistency Across Different Regions]\label{thm:GeneralCaseForFunctionalConsistency}
For the causal model defined in Definition \ref{def:Functional Consistency with respect to different regions}, assume the aggregated variables $\overline{X}_A$, $\overline{X}_B$ and $\overline{Y}_A$, $\overline{Y}_B$ have continuous support. Then, there exists functions $\hat{f}$ and $\hat{g}$ such that $\overline{Y}_A = \hat{f}(\overline{X}_A, N_A)$, $\overline{Y}_B = \hat{f}(\overline{X}_B, N_B)$, $\overline{X}_A = \hat{g}(\overline{Y}_A, N'_A)$, and $\overline{X}_B = \hat{g}(\overline{Y}_B, N'_B)$, where $N_A$ is independent of $\overline{X}_A$, $N_B$ is independent of $\overline{X}_B$, $N'_A$ is independent of $\overline{Y}_A$, and $N'_B$ is independent of $\overline{Y}_B$.
\end{theorem}

\section{Conditional Independence Consistency: Recoverability with Constraint-based Method}\label{section:Conditional Independence Consistency}
Constraint-based causal discovery methods utilize conditional independence to identify the Markov equivalence classes of causal structures. These methods heavily rely on the faithfulness assumption, which posits that every conditional independence in the data corresponds to a d-separation in the underlying causal graph.

The information utilized by constraint-based causal discovery methods is less than that used by FCM-based methods. This implies that the consistency we require for the recoverability of constraint-based methods on temporally aggregated data is less stringent than functional consistency. In essence, we only need the temporally aggregated data to preserve the conditional independence of the summary graph of the underlying causal process.
If the temporal aggregation maintains such conditional independence consistency, then the constraint-based causal discovery method can recover the Markov equivalence class of the summary graph entailed by the underlying true causal process.

\subsection{Definitions and Problem Formulation}

To examine whether the temporally aggregated data preserves the conditional independence consistency with the summary graph of the underlying causal process, we will discuss the three fundamental causal structures of the summary graph: the chain, the fork, and the collider. We will provide theoretical analysis for each of these three fundamental cases respectively.

In subsection \ref{section: functional subsection: def}, we assume the original causal process is VAR(1) and we work with the temporal aggregation of it. But in this section, for analytical convenience we will assume the original model is an aligned version of VAR(1), and work with the temporal aggregation of it. We will show this alignment is reasonable because the temporal aggregation of these two original processes is the same when $k$ is large.

\subsubsection{Aligned Model}
\begin{definition}[Trivariant Aligned Model with Instant Structures]\label{def: Aligned Model with Instant Structures}
The aligned model for VAR model with structure function $f_X,f_Y,f_Z$ incorporating chain-like cross lag effect is given by:

$X_0$, $Y_0$, $Z_0$ are independent and follow the initial distribution.
when $t\geq1$,

\begin{description}
\item[Chain-like Model:] $X_{t}= f_X(X_{t-1}, N_{X,t})$, $Y_{t} = f_Y(X_t, Y_{t-1}, N_{Y,t})$, $Z_{t} = f_Z(Y_t, Z_{t-1}, N_{Z,t})$,
\item[Fork-like Model:] $X_{t}=f_X(X_{t-1},Y_{t}, N_{X,t})$, $Y_{t}=f_Y(Y_{t-1}, N_{Y,t})$, $Z_{t} = f_Z(Y_t, Z_{t-1}, N_{Z,t})$,
\item[Collider-like Model:] $X_{t} = f_X(X_{t-1}, N_{X,t})$, $Y_{t} = f_Y(X_{t},Y_{t-1},Z_{t}, N_{Y,t})$, $Z_{t} = f_Z(Z_{t-1}, N_{Z,t})$,
\end{description}

where $f_X, f_Y, f_Z$ are general functions. $N_{X,t}, N_{Y,t}, N_{Z,t}$ are independent random variables with non-zero variance, which are independent of each other and they are identically distributed and independent across time t.

The temporal summation and aggregation are denoted as $S_X := \sum_{i=1}^k X_i$, $\overline{X} := \frac{S_X}{g(k)}$, and similarly for $S_Y$, $\overline{Y}$, $S_Z$, and $\overline{Z}$. When $k$ is finite, $S_X$, $S_Y$, and $S_Z$ have the same conditionally independent relationship with $\overline{X}$, $\overline{Y}$, and $\overline{Z}$, respectively. Therefore, when discussing conditional independence, we can treat \( S_X \), \( S_Y \), and \( S_Z \) as equivalent to \( \overline{X} \), \( \overline{Y} \), and \( \overline{Z} \).

\end{definition}

\begin{figure}[ht]
\centering
\begin{minipage}{0.32\linewidth}
  \centering
  \includegraphics[width=\linewidth]{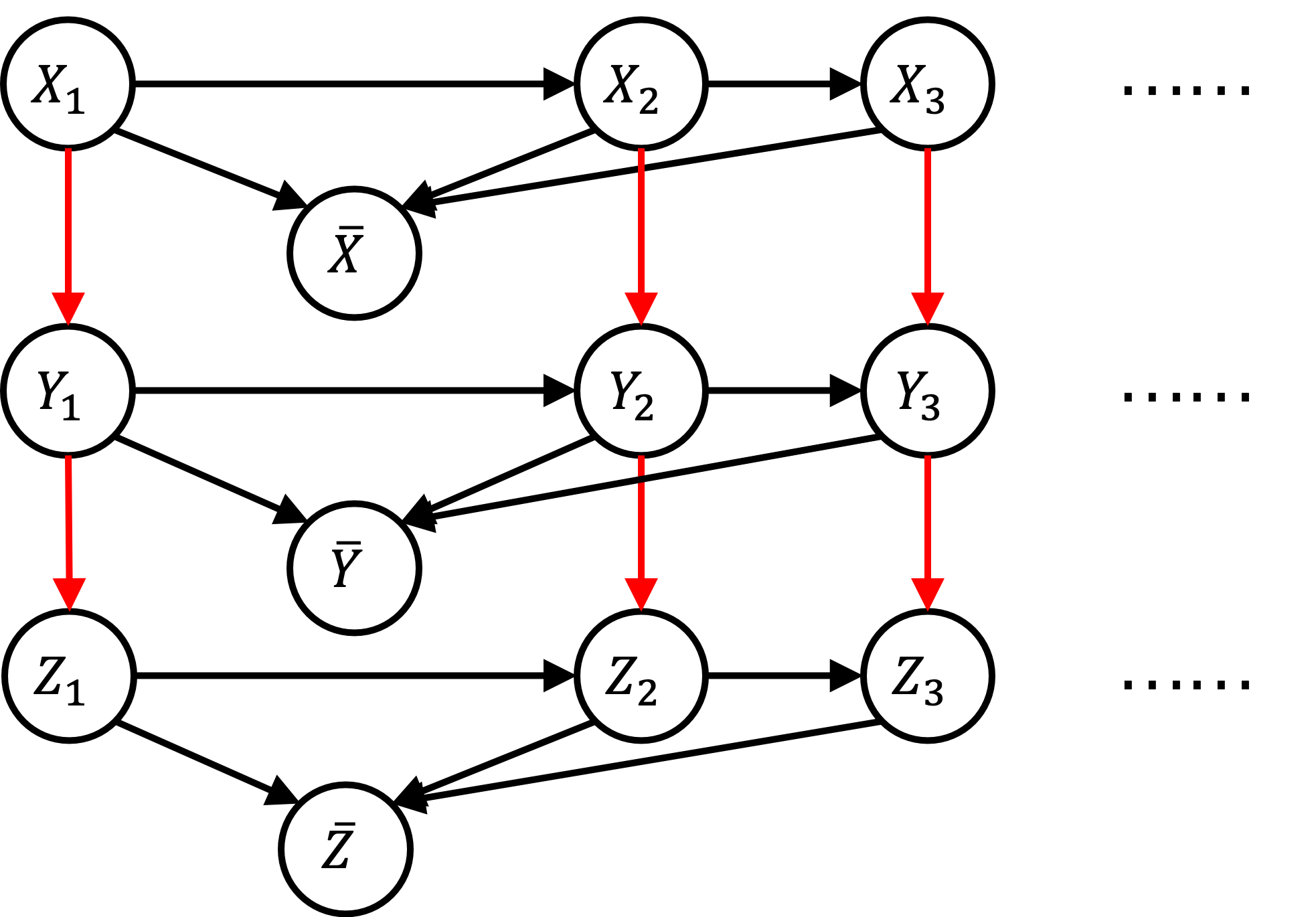}
\end{minipage}%
\hspace{0.01\linewidth}
\begin{minipage}{0.32\linewidth}
  \centering
  \includegraphics[width=\linewidth]{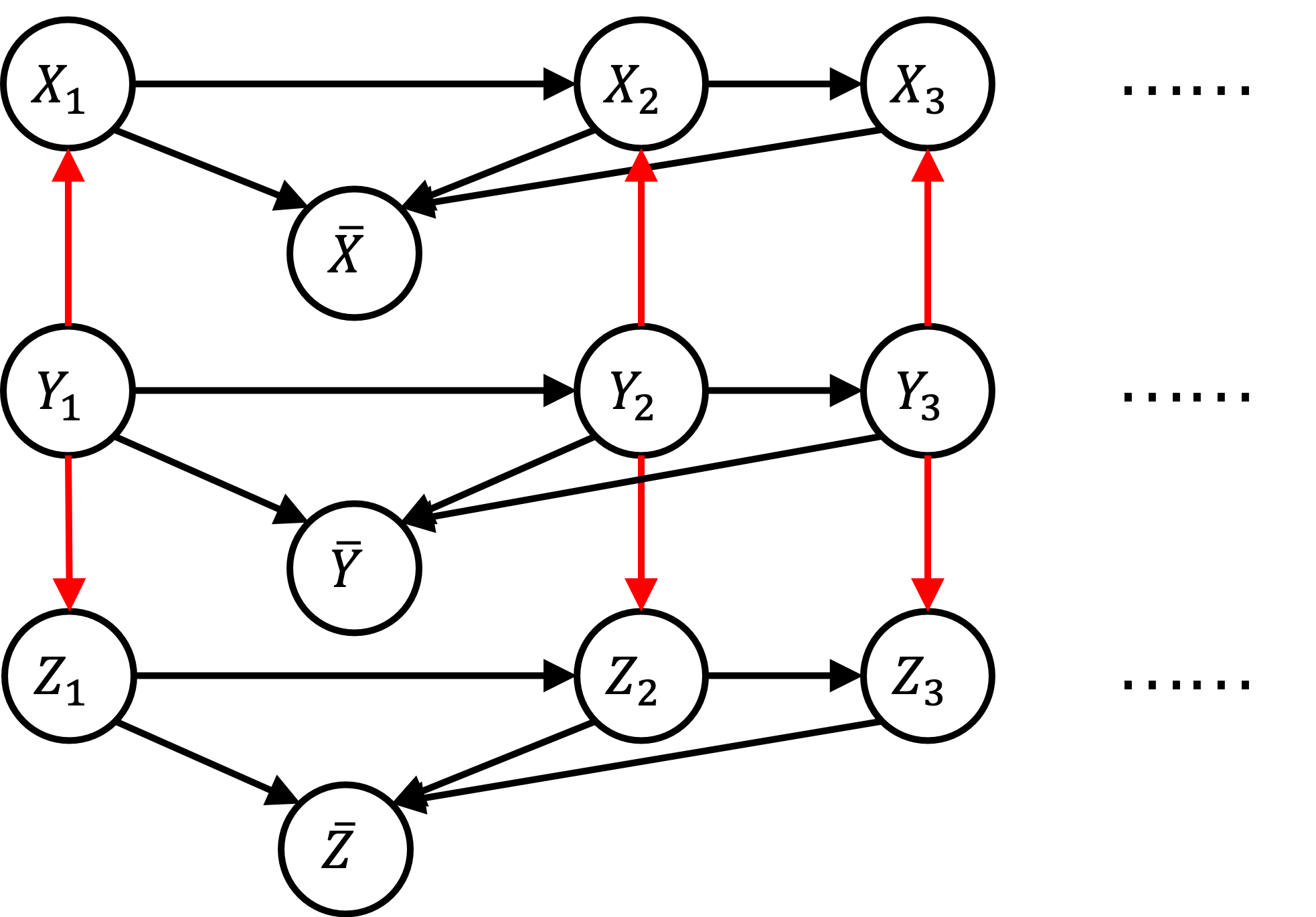}
\end{minipage}%
\hspace{0.01\linewidth}
\begin{minipage}{0.32\linewidth}
  \centering
  \includegraphics[width=\linewidth]{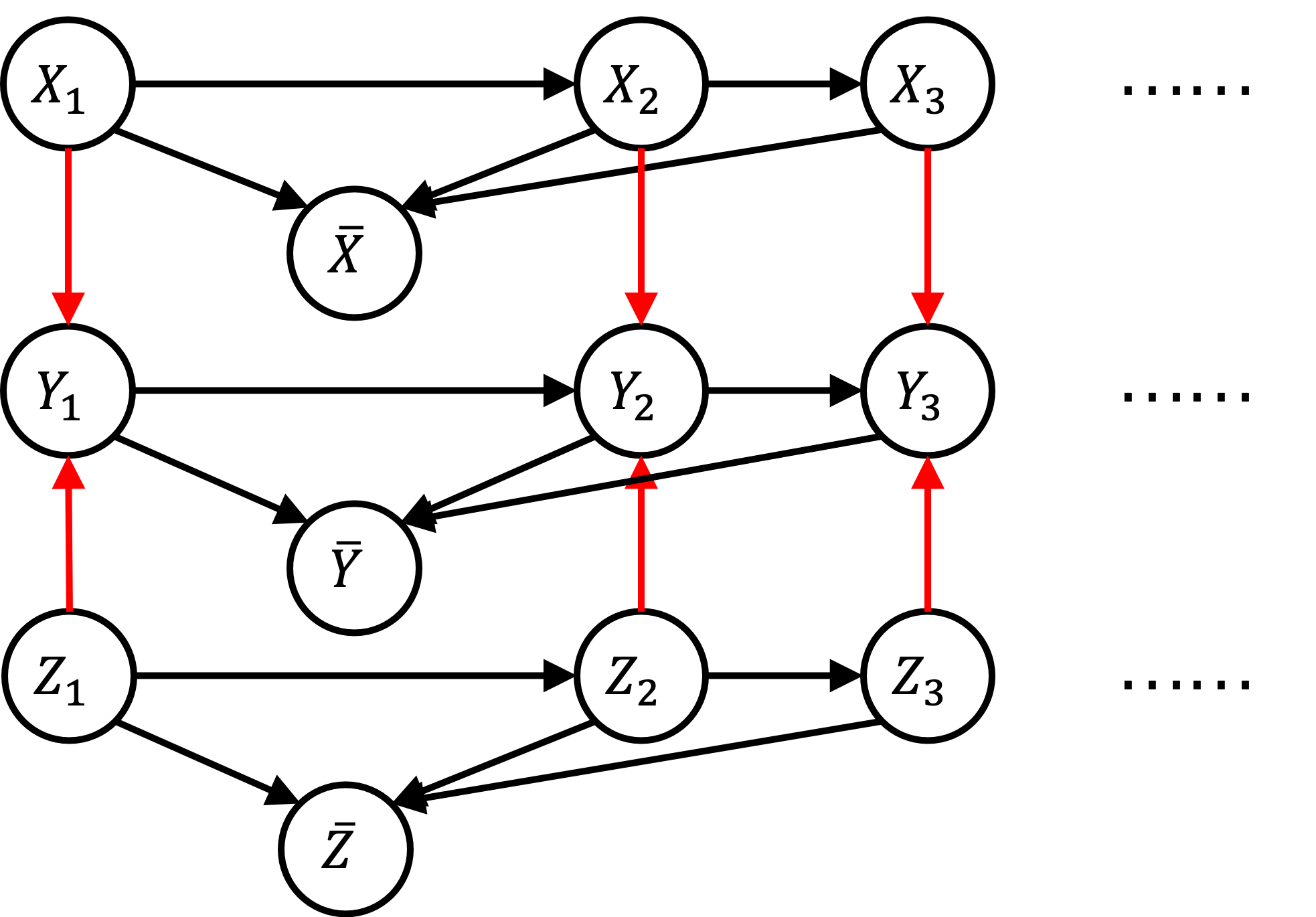}
\end{minipage}

\caption{Left: Chain-like aligned model. Center: Fork-like aligned model. Right: Collider-like aligned model.}
\label{fig:aligned_models} 
\end{figure}

The figures of the aligned models of three fundamental structure involving temporal aggregation variables are presented in Figure \ref{fig:aligned_models}.

\subsubsection{Problem Formulation}

\begin{definition}[Conditional Independence Consistency]
Consider an underlying causal process generating temporally aggregated data. This process is said to exhibit conditional independence consistency if the distribution of temporally aggregated data satisfies the Markov condition with respect to the summary graph of original process.
\end{definition}
We will address the problem of determining the conditions under which temporal aggregation preserves conditional independence consistency in three fundamental causal structures: chain, fork, and collider.

\subsection{Necessary and Sufficient Conditions for Consistency}
For the figure of a chain or fork structure, we expect the middle node can d-separate the nodes on both sides. However, from the structure of Figure \ref{fig:aligned_models} Left and Center, we can find that all adjacent nodes of $\overline{Y}$ point to $\overline{Y}$. Therefore, when we condition on $\overline{Y}$, we cannot block any path. 

For the figure of a collider structure, we expect the nodes on both sides are unconditionally independent and when conditioned on the middle node, the nodes on both sides will be dependent. Fortunately, from the structure of Figure \ref{fig:aligned_models} Right, we can find that all the $Y_t$ are collider for $X_t$ and $Z_t$ so $\overline{X}\perp\!\!\!\perp \overline{Z} $ unconditionally. And because $\overline{Y}$ is a descendant of these colliders, when we condition on $\overline{Y}$, the path involving $X_t$ and $Z_t$ will be open. As a result, $\overline{X}$ is dependent with $\overline{Z}$ conditional on $\overline{Y}$.

\begin{remark}[Conditional Independence Consistency under Faithfulness Condition]\label{remark: chainlike not consistent in general}
Assume the aligned models satisfy the causal Markov condition and causal faithfulness condition with respect to the causal graphs of aligned models in Figure \ref{fig:aligned_models}.
\begin{itemize}
\item The conditional independent sets of temporal aggregation of chain-like/fork-like aligned model is $\emptyset$, which is \textbf{not} consistent with the chain/fork structure.
\item The conditional independent sets of temporal aggregation of collider-like aligned model is ${\overline{X}\perp\!\!\!\perp \overline{Z}}$, which is consistent with the collider structure.
\end{itemize}
\end{remark}

This remark emphasizes that in general cases, the temporal aggregation of a model with chain-/fork-like structure does not exhibit the same conditional independence as a genuine chain or fork structure under the faithfulness assumption. As a result, we will explore the conditions required to ensure the validity of the conditional independence $\overline{X} \perp\!\!\!\perp \overline{Z} \mid \overline{Y}$ in the context of temporal aggregation.

\begin{theorem}[Necessary and Sufficient Condition for Conditional Independence Consistency of Chain and Fork Structure]\label{ci main theorem}
Consider the distribution of ($\overline{X},\overline{Y},\overline{Z},Y_1,...,Y_k$) entailed from the aligned model, the following statements are equivalent when $2 \leq k< \infty$:

\text{(i) Conditional Independence: }  $\overline{X} \perp\!\!\!\perp \overline{Z} \mid \overline{Y}$ \nonumber \\
\text{(ii) Conditional Probability Relation: } $\forall s_{X}, s_{Y}, s_{Z}\in \mathbb{R}$ \nonumber \\
$\iint_{\mathbb{R}^k}$ $\alpha(s_{Z},s_{Y},y_{1:k})\left(\beta(y_{1:k},s_{Y},s_{X})-\gamma(y_{1:k},s_{Y})\right)\,$ \\ $dy_1 ... \,dy_k = 0$\label{z|yx} \\
\text{(iii) Alternative Conditional Probability Relation: } 
\\ $\forall s_{X}, s_{Y}, s_{Z}\in \mathbb{R}$ \nonumber \\
$\iint_{\mathbb{R}^k} \alpha^*(s_{X},s_{Y},y_{1:k})\left(\beta^*(y_{1:k},s_{Y},s_{Z})-\gamma(y_{1:k},s_{Y})\right)\,$\\$dy_1 ... \,dy_k = 0$\label{x|yz}

where
 $\alpha:=p_{S_{Z}|S_{Y},Y_{1:k}}$,
 $\beta:=p_{Y_{1:k}|S_{Y},S_{X}}$,
$\alpha^*:=p_{S_{X}|S_{Y},Y_{1:k}}$,
$\beta^*:=p_{Y_{1:k}|S_{Y},S_{Z}}$,
$\gamma:=p_{Y_{1:k}|S_{Y}}$.

\end{theorem}
See Appendix \ref{proof for main theorem for ci consistency} for the proof. From this sufficient and necessary condition, we find that the integrand can be divided into two parts.
For example, the integrand in formula \ref{z|yx} can be divided into two parts.
The first part is $p_{S_{Z}|S_{Y},Y_{1:k}}(s_{Z}|s_{Y},y_{1:k})$.  Because $Y_1,...Y_k$ d-separate $S_{Z}$ from $X_1,...,X_k$ perfectly, this part is related to the causal mechanism between Y and Z.
The second part $\left(p_{Y_{1:k}|S_{Y},S_{X}}(y_{1:k}|s_{Y},s_{X})-p_{Y_{1:k}|S_{Y}}(y_{1:k}|s_{Y})\right)$ is related to the causal mechanism between Y and Z.
This inspires us to consider different parts of the model individually. 

\begin{corollary}[Sufficient Conditions for Conditional Independence]\label{Sufficient Conditions for Conditional Independence}
    If $\{\overline{X} \perp \!\!\!\perp Y_{1:k}\mid \overline{Y}\}$ \textbf{or} $\{\overline{Z} \perp \!\!\!\perp Y_{1:k}\mid \overline{Y}\}$ holds, then $\overline{X} \perp\!\!\!\perp \overline{Z} \mid \overline{Y}$ holds.
\end{corollary}

Proof can be found in Appendix \ref{proof for sufficient for ci consistency}. This corollary has a very intuitive interpretation: when the information needed to infer $\overline{X}$ from $Y_{1:k}$ is completely included in $\overline{Y}$ ($\overline{X} \perp \!\!\! \perp Y_{1:k} \mid \overline{Y}$), then conditioning on $\overline{Y}$ is equivalent to conditioning on $Y_{1:k}$. In this case, $Y_{1:k}$ d-separate $\overline{X}$ from $\overline{Z}$. The same principle applies to $\overline{Z}$. When does the information to infer $\overline{X}$/$\overline{Z}$ from $Y_{1:k}$ is completely included in $\overline{Y}$?

\begin{corollary}[Partial Linear Conditions]\label{corollary:partial linear is enough}
~
    \begin{enumerate}
        \item For a fork-like aligned model:
        \begin{itemize}
            \item If \(f_Z(Y_t,Z_{t-1},N_{Z,t})\) is of the form \(\alpha*Y_t+N_t\), where \(\alpha\) can be any real number, then \(\overline{Z} \perp \!\!\!\perp Y_{1:k}\mid \overline{Y}\).
            \item If \(f_X(X_{t-1},Y_{t}, N_{X,t})\) is of the form \(\alpha*Y_t+N_t\), where \(\alpha\) can be any real number, then  \(\overline{X} \perp \!\!\!\perp Y_{1:k}\mid \overline{Y}\).
        \end{itemize}
        
        \item For a chain-like aligned model:
        \begin{itemize}
            \item If \(f_Z(Y_t,Z_{t-1},N_{Z,t})\) is of the form \(\alpha*Y_t+N_t\), where \(\alpha\) can be any real number, then \(\overline{Z} \perp \!\!\!\perp Y_{1:k}\mid \overline{Y}\).
            \item If the time series is stationary and Gaussian, and \(f_Y(X_t, Y_{t-1}, N_{Y,t})\) is of the form \(\alpha*X_t+N_t\), where \(\alpha\) can be any real number, then  \(\overline{X} \perp \!\!\!\perp Y_{1:k}\mid \overline{Y}\).
        \end{itemize}
    \end{enumerate}
\end{corollary}

Proof can be found in Appendix \ref{proof for partial linear enough}. Roughly speaking, this corollary suggests that if the causal relationship between X/Z and Y is linear, then the information needed to infer \(\overline{X}\)/\(\overline{Z}\) from \(Y_{1:k}\) is completely included in \(\overline{Y}\). Further, based on the sufficient condition for conditional independence (refer to Corollary \ref{Sufficient Conditions for Conditional Independence}), we can see that it is not necessary for the entire system to be linear. 

\section{Simulation Experiments}\label{section: simulation}
We conducted five experiments to comprehensively address the various aspects of the aggregation problem. Firstly, we applied widely-used causal discovery methods(PC\citep{spirtes2000causation}, FCI\citep{spirtes2013causal}, GES\citep{chickering2002optimal}) to aggregation data with 4 variables, enabling readers to grasp the motivation and core issue discussed in this paper. Secondly and thirdly, we conducted experiments on functional consistency(apply Direct LiNGAM\citep{shimizu2011directlingam}/ANM\citep{hoyer2008nonlinear} to linear/nonlinear data with different aggregation levels) and conditional independence consistency(perform Kernel Conditional Independence test\citep{zhang2012kernel} on aggregated data) to bolster the theorems presented in the main text. Fourthly, we carried out an experiment to investigate the impact of the $k$ value and to justify the approximations made in this paper. Fifthly, we performed the PC algorithm with a skeleton prior on aggregated data and consistently obtained correct results, offering a preliminary solution to the aggregation problem and laying the groundwork for future research in this area. Due to the page limit, we present only a limited number of results in the main text. Detailed settings and results of the five experiments can be found respectively in the Appendices: \ref{causal discovery from aggregated data appendix}, \ref{functional consistency appendix}, \ref{conditional independence sec appendix}, \ref{Effect of k Value appendix}, and \ref{prior knowledge appendix}.

\begin{table*}[h]
\centering
\caption{Rejection rates for CI tests with different combinations of linear and nonlinear relationships. The index in the box represents the conditional independence that the structure should ideally have. Simply speaking, for the column VI, the closer the rejection rate is to 5\%, the better. For all other columns from I to V, a higher rate is better.}
\small
\begin{tabular}{@{}llrrrrrrrr@{}}
\toprule
{$X_{t} \to Y_{t}$} & {$Y_{t} \to Z_{t}$} & {I} & {II} & {III} & {IV} & {V} & {\fbox{VI}} &{A} & {B} \\
\midrule
Linear & Linear & 100\% & 100\% & 100\% & 100\% & 100\% & 5\% & 6\% & 5\%\\
Nonlinear & Linear & 92\% & 100\% & 84\% & 92\% & 100\% & 5\% & 76\% & 5\%\\
Linear & Nonlinear & 100\% & 93\% & 85\% & 100\% & 93\% & 5\% & 5\% & 71\% \\
Nonlinear & Nonlinear & 92\% & 93\% & 72\% & 86\% & 87\% & 58\% & 72\% & 74\%\\
\bottomrule
\end{tabular}
\label{experiment result}
\vspace{-0.3cm}
\end{table*}
\subsection{FCM-based Causal discovery in Linear non-Gaussian case}\label{section:Functional-based Causal discovery in Linear non-Gaussian case}

Here we examine the use of a FCM-based causal discovery method on bivariate temporally aggregated data in the linear non-Gaussian case to distinguish between cause and effect. Specifically, we employ the Direct LiNGAM method to represent FCM-based causal discovery methods.

We perform a simulation experiment on the model $Y_i = 2X_i + N_{Y,i}$, where the noise follow uniform distribution. We then generating the dataset ${\overline{X}, \overline{Y}}$ with a sample size of 10,000. We apply the Direct LiNGAM method on this dataset to determine the causal order.
\begin{wrapfigure}{r}{0.3\textwidth}\vspace{-0.9cm}
\centering
\includegraphics[width=0.99\linewidth]{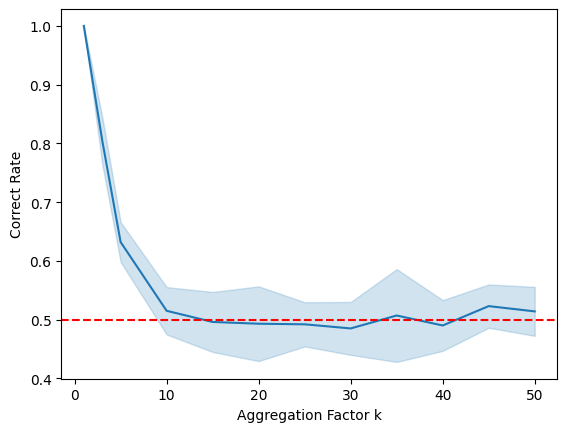}
\caption{Relationship between the aggregation factor $k$ and the performance of the Direct LiNGAM method. The blue area represents the standard deviation. The red line represents the random guess baseline.}
\label{fig:direct-lingam}
\end{wrapfigure}
To investigate the relationship between the aggregation factor $k$ and the performance of the Direct LiNGAM method, we vary the values of $k$ from 1 to 100 to see the accuracy in 100 repetitions.

Our results indicate that when $k$ is small, the accuracy is near 100\%, implying a good performance of the Direct LiNGAM method. However, as $k$ increases from 3 to 30, the accuracy drops rapidly to 50\% as random guess. This experiment demonstrates that even in the linear non-Gaussian case, temporal aggregation can significantly impair the identifiability of functional-based methods relying on non-Gaussianity.

\subsection{Conditional Independence Test in Gaussian case}\label{subsection CI experiment}
We perform experiments using three different structures: chain, fork, and collider to validate our theoretical results.

In all structures, each noise term $N$ follows an independent and identically distributed (i.i.d) standard Gaussian distribution. For the causal relationship between $X$ and $Y$, we use the function $f(\cdot,N)$. In the linear case, $f(\cdot) = (\cdot)$. In the nonlinear case, we use the post-nonlinear model $f(\cdot,N) = G(F(\cdot)+N)$ \citep{zhang2012identifiability} and uniformly randomly pick $F$ and $G$ from $(\cdot)^2$, $(\cdot)^3$, and $tanh(\cdot)$ for each repetition.  This is to ensure that our experiment covers a wide range of nonlinear cases. Similarly, the same approach is applied for the relationship between $Y$ and $Z$ with the corresponding function $g(\cdot)$.

We set $t=1,2$, thus $\overline{X}=X_1+X_2$, $\overline{Y}=Y_1+Y_2$, $\overline{Z}=Z_1+Z_2$. And we generate 1000 i.i.d. data points for $(X_1, X_2, Y_1, Y_2, \overline{X}, \overline{Y})$ and feed them into the approximate kernel-based conditional independence test \citep{strobl2019approximate}.  We test the null hypothesis(conditional independence) for (I) $\overline{X} \perp\!\!\!\perp \overline{Y}$, (II) $\overline{Y} \perp\!\!\!\perp \overline{Z}$, (III) $\overline{X} \perp\!\!\!\perp \overline{Z}$, (IV) $\overline{X} \perp\!\!\!\perp \overline{Y} \mid \overline{Z}$, (V) $\overline{Y} \perp\!\!\!\perp \overline{Z} \mid \overline{X}$, (VI) $\overline{X} \perp\!\!\!\perp \overline{Z} \mid \overline{Y}$. We also test for the conditional independence in corollary \ref{Sufficient Conditions for Conditional Independence}: (A) $\overline{X} \perp\!\!\!\perp Y_1 \mid \overline{Y}$, and (B) $\overline{Z} \perp\!\!\!\perp Y_1 \mid \overline{Y}$. We report the rejection rate for fork structure at a 5\% significance level in 100 repeated experiments in Table \ref{experiment result2}. The results for chain structure and collider structure can be found in Appendix \ref{experiment result1}.

The experiment shows that as long as there is some linearity, we can find a consistent conditional independence set, which aligns with Corollary \ref{corollary:partial linear is enough}. However, in completely non-linear situations, we still cannot find a consistent conditional independence set, which aligns with Remark \ref{remark: chainlike not consistent in general}.

\section{Conclusion and Limitation}
This paper points out that although many people use the occurrence of instantaneous causal relationships due to temporal aggregation as a real-world explanation for instantaneous models, few people pay attention to whether these instantaneous causal relationships are consistent with the underlying time-delayed causal relationships when this situation occurs. This paper mainly discusses whether the causal models generated by temporal aggregation maintain functional consistency and conditional independence consistency in general (nonlinear) situations. Through theoretical analysis in the case of finite $k$, we show that functional consistency is difficult to achieve in non-linear situations. Furthermore, through theoretical analysis and experimental verification in the case of infinite k, we show that even in linear non-Gaussian situations, the instantaneous model generated by temporal aggregation is still unidentifiable. For conditional independence consistency, we show through sufficient and necessary conditions and experiments that it can be satisfied as long as the causal process has some linearity. However, it is still difficult to achieve in completely non-linear situations.

Limitations: Although the negative impact of temporal aggregation on instantaneous causal discovery has been pointed out, a solution has not been provided.

\section*{Acknowledgements}
We would like to acknowledge the support from NSF Grant 2229881, the National Institutes of Health (NIH) under Contract R01HL159805, and grants from Apple Inc., KDDI Research Inc., Quris AI, and Florin Court Capital. MG was supported by ARC DE210101624 and DP240102088.

\section*{Impact Statement}
This study offers a theoretical analysis of potential challenges when applying non-temporal causal discovery algorithms to temporally-aggregated time series data. It provides valuable insights for researchers involved in algorithm design and serves as guidance for practitioners employing causal discovery algorithms. Importantly, the research operates at an abstract level, avoiding any involvement with sensitive data and raising no ethical concerns.


\bibliography{paper}
\bibliographystyle{icml2024}

\newpage
\appendix
\onecolumn


\section{Proof for Necessary Condition of Functional Consistency}\label{functional appendix}

\begin{theorem}[Construction of $\hat{f}$]
If such $\hat{f}$, as defined in Definition \ref{def:FunctionalConsistencyRegardingAdditiveNoise}, exists, then $\hat{f}$ must take the form:
\begin{equation}\label{eq:ConstructionOfHatFappendix}
    \hat{f}(T) = \mathbb{E}\left(\sum_{i=1}^k f(X_i) \mid \overline{X}=T\right) + c,
\end{equation}
where $c$ is any constant (which can be incorporated into the noise term) and the expression \(\mathbb{E}(\cdot \mid \overline{X}=T)\) denotes the conditional expectation. For simplicity, we set $c=0$. Consequently, this implies:
\begin{equation}\label{eq:ExpectationOfHatF}
    \mathbb{E}(\hat{f}(\overline{X})) = \mathbb{E}\left(\sum_{i=1}^k f(X_i)\right).
\end{equation}
\end{theorem}

\begin{proof}\label{proof:additive noise constrction}
According to the definition of functional consistency regarding additive noise, we can express $\overline{Y}$ in two ways: $\overline{Y} = \sum_{i=1}^k f(X_i) + \sum_{i=1}^k N_{Y,i} = \hat{f}(\overline{X}) + N$.

Taking the conditional expectation conditioned on $\overline{X}=T$ for both sides of the equation (noting that both $N_{Y,i}$ and $N$ are independent of $\overline{X}$):
\begin{equation}
\mathbb{E}\left(\sum_{i=1}^k f(X_i) \mid \overline{X}=T\right) + \mathbb{E}\left(\sum_{i=1}^k N_{Y,i}\right) = \hat{f}(T) + \mathbb{E}(N).
\end{equation}

Then, we have:
\begin{equation}
\hat{f}(T)=\mathbb{E}\left(\sum_{i=1}^k f(X_i) \mid \overline{X}=T\right) + \left(\mathbb{E}\left(\sum_{i=1}^k N_{Y,i}\right)-\mathbb{E}(N)\right).
\end{equation}

Please note that $\mathbb{E}\left(\sum_{i=1}^k N_{Y,i}\right)-\mathbb{E}(N)$ is a constant, not related to $T$.

Therefore, 
\begin{equation}
\hat{f}(T)=\mathbb{E}\left(\sum_{i=1}^k f(X_i) \mid \overline{X}=T\right) + c.
\end{equation}

Finally, applying the law of total expectation, we obtain the expectation of \(\hat{f}\).
\end{proof}

\begin{theorem}[Necessary and Sufficient Condition]
The necessary and sufficient condition for the existence of the additive noise causal model defined in Definition \ref{def:FunctionalConsistencyRegardingAdditiveNoise} is that $N = \sum_{i=1}^k N_{Y,i} + \left( \sum_{i=1}^k f(X_i) - \hat{f}(\overline{X}) \right)$ is independent of $\overline{X}$,
where $\hat{f}$ is defined by Eq. \ref{eq:ConstructionOfHatF}.
\end{theorem}
\begin{proof}\label{proof for necessary and sufficient condition for additive}
According to Definition \ref{def:Bivariate Aligned Model with Instant Structures}, $\overline{Y}$ is described as
\begin{equation}
\overline{Y} = \sum_{i=1}^k f(X_i) + \sum_{i=1}^k N_{Y,i}. \label{eq:AdditiveFor1}
\end{equation}

Furthermore, according to Theorem \ref{thm:ConstructionOfHatF}, $\overline{Y}$ can be expressed as
\begin{equation}
\overline{Y} = \hat{f}(\overline{X}) + N. \label{eq:AdditiveFor2}
\end{equation}

By combining Equations \ref{eq:AdditiveFor1} and \ref{eq:AdditiveFor2}, we derive the expression for \(N\) as
\begin{equation}
N = \sum_{i=1}^k N_{Y,i} + \left( \sum_{i=1}^k f(X_i) - \hat{f}(\overline{X}) \right).
\end{equation}

Given the requirement in Definition \ref{def:Bivariate Aligned Model with Instant Structures} that \(N\) should be independent of \(\overline{X}\), we thereby establish the theorem.

\end{proof}

\begin{theorem}[General Case for Functional Consistency Across Different Regions]
For the causal model defined in Definition \ref{def:Functional Consistency with respect to different regions}, assume the aggregated variables $\overline{X}_A$, $\overline{X}_B$ and $\overline{Y}_A$, $\overline{Y}_B$ have continuous support. Then, there exists functions $\hat{f}$ and $\hat{g}$ such that $\overline{Y}_A = \hat{f}(\overline{X}_A, N_A)$, $\overline{Y}_B = \hat{f}(\overline{X}_B, N_B)$, $\overline{X}_A = \hat{g}(\overline{Y}_A, N'_A)$, and $\overline{X}_B = \hat{g}(\overline{Y}_B, N'_B)$, where $N_A$ is independent of $\overline{X}_A$, $N_B$ is independent of $\overline{X}_B$, $N'_A$ is independent of $\overline{Y}_A$, and $N'_B$ is independent of $\overline{Y}_B$.
\end{theorem}

\begin{proof}\label{proof for diffrent regions}

We only provide the proof for $\overline{Y}_A = \hat{f}(\overline{X}_A, N_A)$, $\overline{Y}_B = \hat{f}(\overline{X}_B, N_B)$ here. And the construction of $\hat{g}$ is very similar to $\hat{f}$.

Aligned with Lemma 1 in ~\citep{zhang2015estimation}, for any random variables, we can construct independent noise with $N_A=q\circ F_{\overline{Y}_A|\overline{X}_A}(\overline{Y}_A)$ and $N_B=F_{\overline{Y}_B|\overline{X}_B}(\overline{Y}_B)$. Since $q$ is an arbitrary, strictly increasing function and the support of the random variable is continuous, the CDF is invertible. We can set $q=F_{\overline{Y}_B|\overline{X}_B}\circ F^{-1}_{\overline{Y}_A|\overline{X}_A}$, making $N_A$ constructed by the same function as $N_B$. Given the CDF's invertibility, we can define $\hat{f}:=F_{\overline{Y}_B|\overline{X}_B}^{-1}$ to derive $\overline{Y}_A = \hat{f}(\overline{X}_A, N_A)$, $\overline{Y}_B = \hat{f}(\overline{X}_B, N_B)$.
    
\end{proof}

\section{Proofs for Conditions of Conditional Independence Consistency}\label{conditional independence appendix}

\begin{theorem}[Necessary and Sufficient Condition for Conditional Independence Consistency]
Consider the distribution of ($\overline{X},\overline{Y},\overline{Z},Y_1,...,Y_k$) entailed from the aligned model, the following statements are equivalent when $2 \leq k< \infty$:
\begin{align}
&\text{(i) Conditional Independence: }  \overline{X} \perp\!\!\!\perp \overline{Z} \mid \overline{Y} \nonumber \\
&\text{(ii) Conditional Probability Relation: } \forall s_{X}, s_{Y}, s_{Z}\in \mathbb{R} \nonumber \\
&\iint_{\mathbb{R}^k} \alpha(s_{Z},s_{Y},y_{1:k})\left(\beta(y_{1:k},s_{Y},s_{X})-\gamma(y_{1:k},s_{Y})\right)\,dy_1 ... \,dy_k = 0 \\
&\text{(iii) Alternative Conditional Probability Relation: } \forall s_{X}, s_{Y}, s_{Z}\in \mathbb{R} \nonumber \\
&\iint_{\mathbb{R}^k} \alpha^*(s_{X},s_{Y},y_{1:k})\left(\beta^*(y_{1:k},s_{Y},s_{Z})-\gamma(y_{1:k},s_{Y})\right)\,dy_1 ... \,dy_k = 0
\end{align}

where
\begin{itemize}
  \item $\alpha(s_{Z},s_{Y},y_{1:k}):=p_{S_{Z}|S_{Y},Y_{1:k}}(s_{Z}|s_{Y},y_{1:k})$
  \item $\beta(y_{1:k},s_{Y},s_{X}):=p_{Y_{1:k}|S_{Y},S_{X}}(y_{1:k}|s_{Y},s_{X})$
  \item $\alpha^*(s_{X},s_{Y},y_{1:k}):=p_{S_{X}|S_{Y},Y_{1:k}}(s_{X}|s_{Y},y_{1:k})$
  \item $\beta^*(y_{1:k},s_{Y},s_{Z}):=p_{Y_{1:k}|S_{Y},S_{Z}}(y_{1:k}|s_{Y},s_{Z})$
  \item $\gamma(y_{1:k},s_{Y}):=p_{Y_{1:k}|S_{Y}}(y_{1:k}|s_{Y})$
\end{itemize}
\end{theorem}

\begin{proof}\label{proof for main theorem for ci consistency}

The proof will proceed by showing that statements (i), (ii), and (iii) are mutually equivalent. 

\textit{Proof that (i) is equivalent to (ii)}: Suppose that $S_{X} \perp\!\!\!\perp S_{Z} \mid S_{Y}$ holds. By the definition of conditional independence, this is equivalent to the statement that for all $s_{X}$, $s_{Y}$, and $s_{Z}$ in $\mathbb{R}$, we have $p_{S_{Z}|S_{Y},S_{X}}(s_{Z}|s_{Y},s_{X}) = p_{S_{Z}|S_{Y}}(s_{Z}|s_{Y})$.

We can now derive both sides of this equation as follows. 

On the left hand side(LHS):
\begin{align*}
&p_{S_{Z}|S_{Y},S_{X}}(s_{Z}|s_{Y},s_{X})\\
&= \frac{p_{S_{Z},S_{Y},S_{X}}(s_{Z},s_{Y},s_{X})}{p_{S_{Y},S_{X}}(s_{Y},s_{X})} \nonumber \\
&= \frac{\iint_{\mathbb{R}^k} p_{S_{Z}|S_{Y},S_{X},Y_{1:k}}(s_{Z}|s_{Y},s_{X},y_{1:k})p_{S_{Y},S_{X},Y_{1:k}}(s_{Y},s_{X},y_{1:k})dy_1 ... dy_k}{p_{S_{Y},S_{X}}(s_{Y},s_{X})} \nonumber \\
&= \iint_{\mathbb{R}^k} p_{S_{Z}|S_{Y},Y_{1:k}}(s_{Z}|s_{Y},y_{1:k})p_{Y_{1:k}|S_{Y},S_{X}}(y_{1:k}|s_{Y},s_{X})dy_1 ... dy_k \nonumber \\
&= \iint_{\mathbb{R}^k} \alpha(s_{Z},s_{Y},y_{1:k})\beta(y_{1:k},s_{Y},s_{X})dy_1 ... dy_k \nonumber 
\end{align*}
The first steps is based on the definition of conditional probability and the second step uses the law of total probability. The third step is using the d-separation: $\{Y_1, \dots, Y_k\}$ d-separates $S_{Z}$ from $S_{X}$.

Meanwhile, RHS:
\begin{align*}
    &p_{S_{Z}|S_{Y}}(s_{Z}|s_{Y}) \\
    &= \frac{p_{S_{Z},S_{Y}}(s_{Z},s_{Y})}{p_{S_{Y}}(s_{Y})} \nonumber \\
    &= \frac{\iint_{\mathbb{R}^k} p_{S_{Z}|S_{Y},Y_{1:k}}(s_{Z}|s_{Y},y_{1:k})p_{S_{Y},Y_{1:k}}(s_{Y},y_{1:k})dy_1 ... dy_k}{p_{S_{Y}}(s_{Y})} \\
    &=\iint_{\mathbb{R}^k} \alpha(s_{Z}|s_{Y},y_{1:k})\gamma(y_{1:k},s_{Y})dy_1 ... dy_k
\end{align*}

Finally, substitute both to the original equality:
\begin{align*}
    &p_{S_{Z}|S_{Y},S_{X}}(s_{Z}|s_{Y},s_{X})-p_{S_{Z}|S_{Y}}(s_{Z}|s_{Y})\\
    &=\iint_{\mathbb{R}^k} \alpha(s_{X},s_{Y},y_{1:k})\left(\beta(y_{1:k},s_{Y},s_{Z})-\gamma(y_{1:k},s_{Y})\right)\,dy_1 ... \,dy_k \\
    &=0
\end{align*}

Hence, we arrive at the condition specified in (ii).

\textit{Proof that (i) is equivalent to (iii)}: The proof that (i) and (iii) are equivalent is analogous to the above arguments. We therefore omit the details for brevity.

\end{proof}

\begin{corollary*}[Sufficient Conditions for Conditional Independence]
    If $\{\overline{X} \perp \!\!\!\perp Y_{1:k}\mid \overline{Y}\}$ \textbf{or} $\{\overline{Z} \perp \!\!\!\perp Y_{1:k}\mid \overline{Y}\}$ hold, then $\overline{X} \perp\!\!\!\perp \overline{Z} \mid \overline{Y}$ holds.
\end{corollary*}
\begin{proof}\label{proof for sufficient for ci consistency}
This corollary introduces two sufficient conditions for conditional independence. While the proofs for each are analogous, we demonstrate the proof for the first condition to avoid redundancy. 

\emph{Proof that $\{\overline{X} \perp \!\!\!\perp Y_{1:k}\mid \overline{Y}\}$ implies $\overline{X} \perp\!\!\!\perp \overline{Z} \mid \overline{Y}$}:  

Assume that $\{\overline{X} \perp \!\!\!\perp Y_{1:k}\mid \overline{Y}\}$ is true. By definition, this is equivalent to 

\begin{align*}
    p_{Y_{1:k}|S_{Y},S_{X}}(y_{1:k}|s_{Y},s_{X})=p_{Y_{1:k}|S_{Y}}(y_{1:k}|s_{Y})
\end{align*}

for all $s_X$, $y_{1:k}$, and $s_Y$. Utilizing the notation from Theorem \ref{ci main theorem}, we can rewrite this as

\begin{align*}
    \beta(y_{1:k},s_{Y},s_{X})=\gamma(y_{1:k},s_{Y}).
\end{align*}

This simplification makes it clear that the equality conforms to the second condition of Theorem \ref{ci main theorem}, which is 

\begin{align*}
    \iint_{\mathbb{R}^k} \alpha(s_{Z},s_{Y},y_{1:k})\left(\beta(y_{1:k},s_{Y},s_{X})-\gamma(y_{1:k},s_{Y})\right)\,dy_1 ... \,dy_k = 0.
\end{align*}

Because this holds for all $s_X$, $s_Y$, and $s_Z$ in $\mathbb{R}$, it follows that $\overline{X} \perp\!\!\!\perp \overline{Z} \mid \overline{Y}$ is true, completing our proof. 

\end{proof}

\begin{corollary*}
    \begin{enumerate}
        \item For a fork-like aligned model:
        \begin{itemize}
            \item If \(f_Z(Y_t,Z_{t-1},N_{Z,t})\) is of the form \(\alpha*Y_t+N_t\), where \(\alpha\) can be any real number, then \(\overline{Z} \perp \!\!\!\perp Y_{1:k}\mid \overline{Y}\).
            \item If \(f_X(X_{t-1},Y_{t}, N_{X,t})\) is of the form \(\alpha*Y_t+N_t\), where \(\alpha\) can be any real number, then  \(\overline{X} \perp \!\!\!\perp Y_{1:k}\mid \overline{Y}\).
        \end{itemize}
        
        \item For a chain-like aligned model:
        \begin{itemize}
            \item If \(f_Z(Y_t,Z_{t-1},N_{Z,t})\) is of the form \(\alpha*Y_t+N_t\), where \(\alpha\) can be any real number, then \(\overline{Z} \perp \!\!\!\perp Y_{1:k}\mid \overline{Y}\).
            \item If the time series is stationary and Gaussian, and \(f_Y(X_t, Y_{t-1}, N_{Y,t})\) is of the form \(\alpha*X_t+N_t\), where \(\alpha\) can be any real number, then  \(\overline{X} \perp \!\!\!\perp Y_{1:k}\mid \overline{Y}\).
        \end{itemize}
    \end{enumerate}
\end{corollary*}

\begin{proof}\label{proof for partial linear enough}
The proof for these four sufficient conditions is tied to the bivariate substructure within the fork and chain models.

We categorize the bivariate substructures within these trivariate structures into two types. The first type is where the middle node directs the side nodes, such as in the fork structure where the middle node \(Y\) directs \(X\) and \(Z\). There are two such substructures in the fork model and one in the chain model where \(Y\) directs \(Z\). The second type is where the side node directs the middle node, seen in the chain model with \(X\) directing \(Y\).

Due to the causal direction in the bivariate structure, the sufficient conditions for \(\overline{Z} \perp \!\!\!\perp Y_{1:k}\mid \overline{Y}\) and \(\overline{X} \perp \!\!\!\perp Y_{1:k}\mid \overline{Y}\) in the fork, and \(\overline{Z} \perp \!\!\!\perp Y_{1:k}\mid \overline{Y}\) in the chain are similar and share a similar proof. The sufficient condition for \(\overline{X} \perp \!\!\!\perp Y_{1:k}\mid \overline{Y}\) in the chain is different from the other three. To avoid redundancy, we provide a proof for the sufficient condition for \(\overline{Z} \perp \!\!\!\perp Y_{1:k}\mid \overline{Y}\) in the chain; the proof for the two conditions in the fork model is similar to this. We also provide the proof for the sufficient condition for \(\overline{X} \perp \!\!\!\perp Y_{1:k}\mid \overline{Y}\) in the chain.

\textit{proof for sufficient condition of $\overline{Z} \perp \!\!\!\perp Y_{1:k}\mid \overline{Y}$ in chain model:}

Suppose \(f_Z(Y_t,N_{Z,t})=\alpha*Y_t+N_{Z,t}\)  for some real number $alpha$. Then, by substitution, we get 

\begin{align*}
    S_Z&=\sum_{t=1}^k Z_t\\
    &=\sum_{t=1}^k (\alpha Y_t+N_{Z,t})\\
    &=\alpha S_Y+\sum_{t=1}^k N_{Z,t}
\end{align*}
Given $S_Y$, the random part of $S_Z$ is only $\sum_{t=1}^k N_{Z,t}$, which is independent of $Y_{1:k}$. Therefore, it follows that $\overline{Z} \perp \!\!\!\perp Y_{1:k}\mid \overline{Y}$.

\textit{proof for sufficient condition of $\overline{X} \perp \!\!\!\perp Y_{1:k}\mid \overline{Y}$ in chain model:}

We will prove the case for $k = 2$, and it can be easily generalized to $k \geq 3$. In the linear, Gaussian, stationary model for $k = 2$, we have:

\begin{align*}
X_1 &\sim \mathcal{N}(0, \sigma_{X_1}^2) \\
Y_1 &= \alpha X_1 + N_{Y,1} \\
X_2 &= \beta X_1 + N_{X,2} \\
Y_2 &= \alpha X_2 + N_{Y,2}
\end{align*}

where $N_{X,2} \sim \mathcal{N}(0, \sigma_{N_{X}}^2)$, $N_{Y,1}, N_{Y,2} \text{ i.i.d.} \sim \mathcal{N}(0, \sigma_{N_{Y}}^2)$. And due to stationarity, $\sigma_{X_1}^2 = \frac{\sigma_{N_{X}}^2}{1 - \beta^2}$.

In the linear Gaussian case, conditional independence implies that the partial correlation equals 0. We have:

\begin{align*}
\mathrm{cov}_{S_{X},Y_1 \mid S_{Y}} &= \mathrm{cov}_{S_X,Y_1} - \frac{\mathrm{cov}_{S_{X},S_{Y}} \mathrm{cov}_{Y_1,S_{Y}}}{\mathrm{var}_{S_{Y}}} \\
\mathrm{cov}_{S_{X},Y_2 \mid S_{Y}} &= \mathrm{cov}_{S_{X},Y_2} - \frac{\mathrm{cov}_{S_{X},S_{Y}} \mathrm{cov}_{Y_2,S_{Y}}}{\mathrm{var}_{S_{Y}}}
\end{align*}

where 
\begin{align*}
\mathrm{cov}(S_{X}, Y_1) &= \mathrm{cov}(X_1, Y_1) + \mathrm{cov}(X_2, Y_1), \\
\mathrm{cov}(S_{X}, Y_2) &= \mathrm{cov}(X_1, Y_2) + \mathrm{cov}(X_2, Y_2), \\
\mathrm{cov}(S_{X}, S_{Y}) &= \mathrm{cov}(S_{X}, Y_1) + \mathrm{cov}(S_{X}, Y_2), \\
\mathrm{cov}(Y_1, S_{Y}) &= \mathrm{var}(Y_1) + \mathrm{cov}(Y_1, Y_2), \\
\mathrm{cov}(Y_2, S_{Y}) &= \mathrm{var}(Y_2) + \mathrm{cov}(Y_1, Y_2).
\end{align*}

Substitute these into the partial covariance equations to get

$\mathrm{cov}_{S_{X}, Y_1 \cdot Z}=\mathrm{cov}_{S_{X}, Y_2 \cdot Z}=0$

\end{proof}

\section{Causal Discovery from Aggregated Data}\label{causal discovery from aggregated data appendix}

To investigate the direct effects of aggregation on causal discovery, we applied three widely-used causal discovery methods on both the original and aggregated data, comparing the results in both linear and nonlinear scenarios. The performance of these methods is measured using the accuracy over 100 repetitions.

\begin{figure}[H]
\centering
\includegraphics[width=0.3\textwidth]{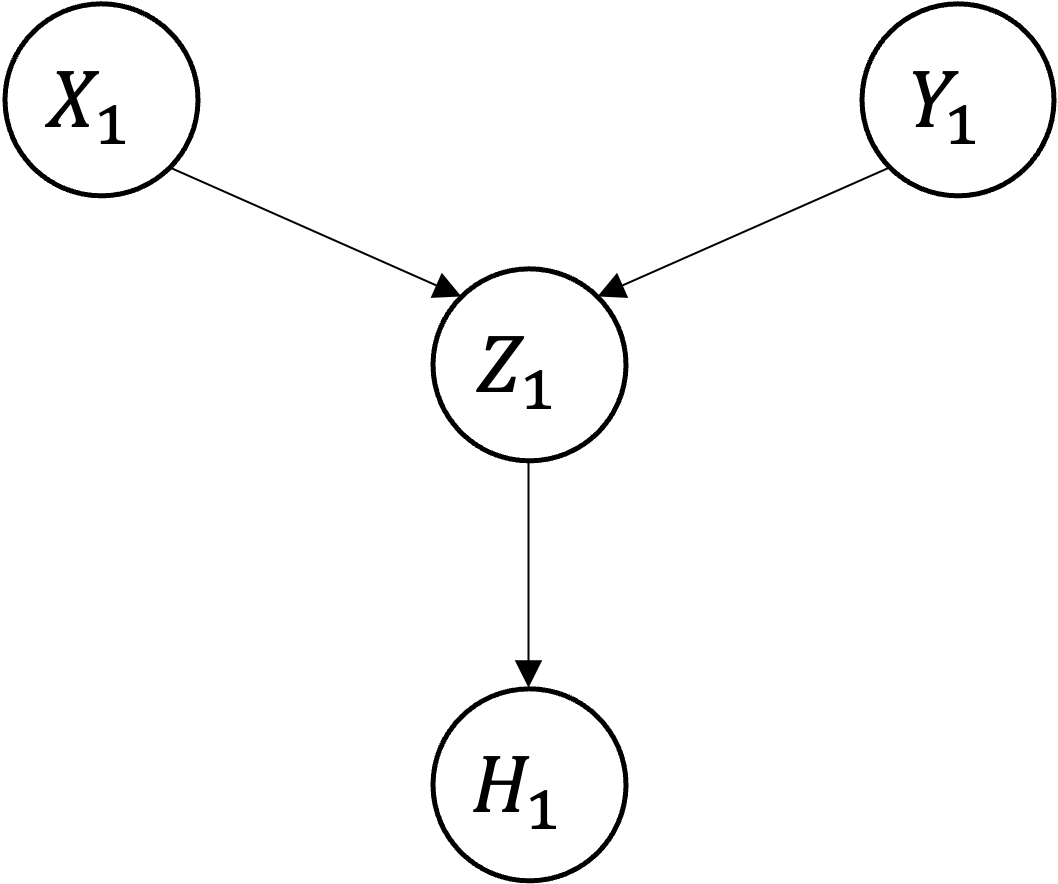}
\caption{Causal graph of original data}
\label{fig:Causal graph of original data}
\end{figure}

For data generation, in each repetition, the original data is generated based on the causal graph shown in Figure \ref{fig:Causal graph of original data}. The causal relationships are defined as:
\begin{align*}
Z &= X + Y + N_Z, \\
H &= Z + N_H \quad \text{(for linear)}; \\
Z &= X^2 + Y^2 + N_Z, \\
H &= Z^2 + N_H \quad \text{(for nonlinear)}.
\end{align*}
The aggregated data is the result of aggregation with a factor of \(k=2\) based on the aligned model \ref{def: Aligned Model with Instant Structures}, having an instant structure resembling the original data. All datasets have a sample size of 500.

Regarding the method parameters, we used the Fisher-Z test for PC and FCI, and the BIC score for GES in the linear scenario. In the nonlinear scenario, we set the conditional independence test for PC and FCI as the Kernel Conditional Independence Test (KCI) with the default kernel and chose the ``local score CV general'' score function for GES. All other settings are kept at their default values.

\begin{figure}[H]
\centering
\includegraphics[width=0.8\textwidth]{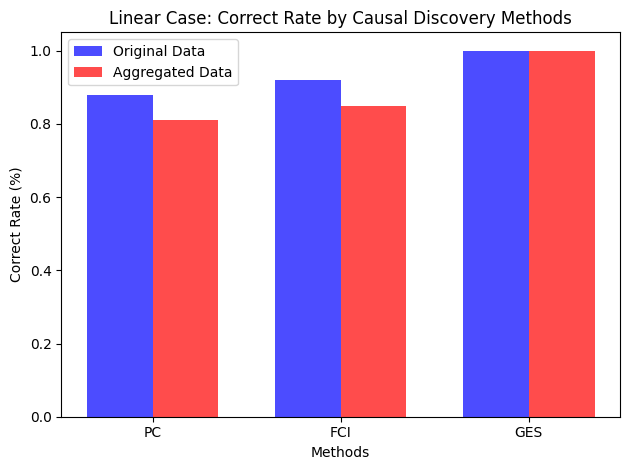}
\caption{Linear Case: Accuracy of Causal Discovery Method}
\label{fig:Linear Case: Correction Rate by Causal Discovery Method}
\end{figure}

From Figure \ref{fig:Linear Case: Correction Rate by Causal Discovery Method}, we observe that, in the linear scenario, aggregation does not adversely affect the performance of the causal discovery methods. This might explain why the causal community has not prioritized the aggregation issue in instantaneous causal discovery for a long time.

\begin{figure}[H]
\centering
\includegraphics[width=0.8\textwidth]{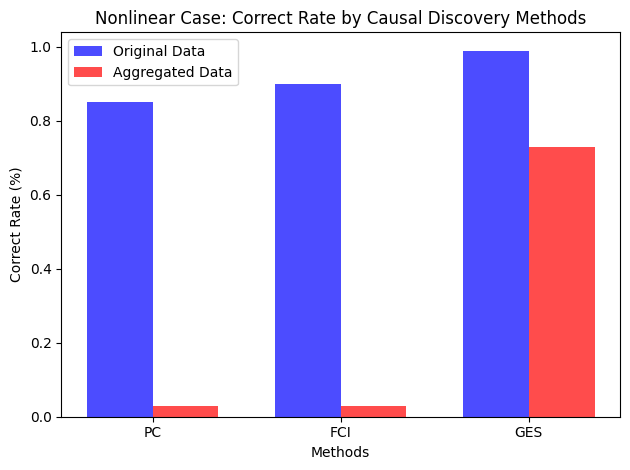}
\caption{Nonlinear Case: Accuracy of Causal Discovery Method}
\label{fig:Nonlinear Case: Correction Rate by Causal Discovery Method}
\end{figure}

Contrastingly, the nonlinear scenario paints a completely different picture, with aggregation causing a significant drop in the performance of all three methods. It is crucial to rigorously investigate this issue to understand its causes and potential solutions.

Additionally, Figure \ref{fig:Nonlinear Case: Correction Rate by Causal Discovery Method} shows the comparison of experimental results of PC, FCI, and GES on nonlinear disaggregated/aggregated data. All three methods have near 90\% accuracy for disaggregated data but experience a significant collapse when they come to aggregated data. The GES method is least affected, still maintaining an accuracy of around 70\%. Therefore, we conducted more experiments to test whether this is due to coincidence or because GES truly has the ability to handle aggregated data.

Motivated by \citet{reisach2021beware}, some score-based methods are sensitive to variance, which may result in falsely high performance. So, we changed the variance of some independent noise in the underlying model:

\[
\begin{aligned}
    &\text{Underlying model: } X_t = N_{X,t}, \; Y_t = N_{Y,t}, \; Z_t = X^2_t + Y^2_t + N_{Z,t}, \; H_t = Z^2_t + N_{H,t}.\\
    &\text{All the independent noises } N \text{ iid follow } N(0,1). \\
    &\text{We respectively changed } N_{X,t} \text{ and } N_{Z,t} \text{ to } N(0,4), \text{ keeping other settings unchanged to perform experiments.}
\end{aligned}
\]

\begin{table}[H]
\centering
\begin{tabular}{@{}lcc@{}}
\toprule
Experimental Change & Disaggregated Accuracy (50 reps) & Aggregated Accuracy \\ \midrule
Increase VAR($N_{X,t}$) & 0.98 $\pm$ 0.03 & 0.13 $\pm$ 0.04 \\
Increase VAR($N_{Z,t}$) & 0.94 $\pm$ 0.04 & 0.28 $\pm$ 0.09 \\ \bottomrule
\end{tabular}
\caption{The accuracy was calculated by repeating the experiment 50 times, and in the outside loop, we repeated this process 5 times to calculate the standard deviation for the accuracy. The results were rounded to two decimal places.}
\end{table}

From the results, we find that even when we increase the variance of one of the independent noises, GES performs very well on disaggregated data. However, when it comes to aggregated data, the accuracy collapses to lower than 30\%.

Therefore, there is no solid evidence to confirm that GES has the ability to deal with aggregated data, but it does have less collapse than other methods in some cases, and it may need further investigation in future work.

\section{Experiment for Functional Consistency}\label{functional consistency appendix}

Determining the causal direction between two variables is an essential task in causal discovery. To understand the impact of aggregation on this task, we employed two renowned FCM-based causal discovery methods: Direct LiNGAM for the linear scenario and Additive Nonlinear Model (ANM) for the nonlinear one. We assessed how the accuracy in 100 repetitions varies with the aggregation factor \(k\).

For data generation, it's straightforward. The data is generated based on an aligned model with an instantaneous causal relationship, given by:
\begin{align*}
Y &= 2X + N_Y \quad \text{(for linear)}; \\
Y &= X^2 + N_Y \quad \text{(for nonlinear)},
\end{align*}
where the independent noise follows a standard uniform distribution. The sample size is 500.

\begin{figure}[H]
\centering
\includegraphics[width=0.8\textwidth]{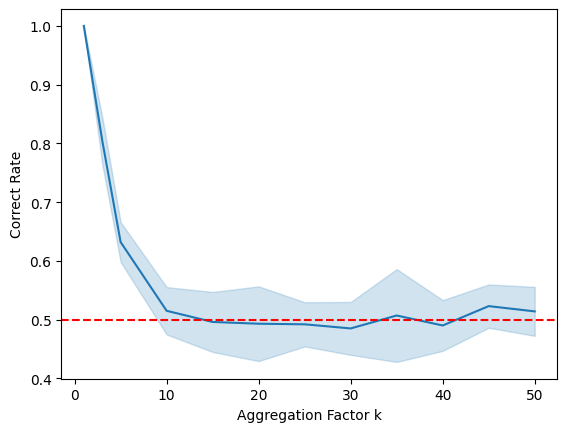}
\caption{Linear Case: Direct LiNGAM Accuracy with Different Aggregation Factors \(k\). The blue area represents the standard deviation. The red line represents the random guess baseline.}
\label{fig:Linear Case: Direct LiNGAM Correction Rate in different aggregation factor k}
\end{figure}

\begin{figure}[H]
\centering
\includegraphics[width=0.8\textwidth]{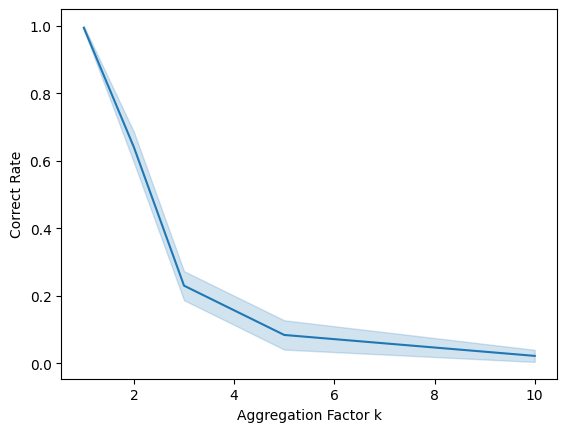}
\caption{Nonlinear Case: ANM Accuracy with Different Aggregation Factors \(k\). The blue area represents the standard deviation.}
\label{fig:Nonlinear Case: ANM Correction Rate in different aggregation factor k}
\end{figure}

From the presented figures, it's evident that in the linear non-Gaussian case, the non-Gaussian distribution increasingly approaches a Gaussian one as \(k\) grows. Eventually, Direct LiNGAM resembles a random guess. It's noteworthy that the x-axis range for the linear scenario spans from 0 to 50, while for the nonlinear case, it's from 0 to 10. This difference suggests that the performance of ANM deteriorates faster than Direct LiNGAM. ANM is not reliant on non-Gaussian properties but on additive noise. Aggregated data lacks functional consistency as the additive noise function is disrupted by the aggregation, rendering ANM ineffective.

\section{Experiment for Conditional Independence Consistency}\label{conditional independence sec appendix}

We conduct experiments using three different structures: chain, fork, and collider, to validate our theoretical results in Section \ref{section:Conditional Independence Consistency}. However, due to space constraints, we only report the results for the fork-like model in the main text. In this section, we will reiterate the experiment details and report the complete results for chain, fork, and collider structures, along with a more detailed analysis.

The specific settings for these structures are as follows:

\begin{description}
\item[Chain-like Model:] $X_t = N_{X,t}$ , $Y_t = f(X_t, N_{Y,t})$ , $Z_t = g(Y_t,N_{Z,t})$
\item[Fork-like Model:] $X_t = f(Y_t, N_{X,t})$, $Y_t = N_{Y,t}$, $Z_t = g(Y_t,N_{Z,t})$
\item[Collider-like Model:] $X_t = N_{X,t}$ , $Y_t = f(X_t,N_{Y,t}) + g(Z_t, N_{Y,t})$ , $Z_t = N_{Z,t}$
\end{description}

In all structures, each noise term $N$ follows an independent and identically distributed (i.i.d) standard Gaussian distribution. For the causal relationship between $X$ and $Y$, we use the function $f(\cdot,N)$. In the linear case, $f(\cdot) = (\cdot)$. In the nonlinear case, we use the post-nonlinear model $f(\cdot,N) = G(F(\cdot)+N)$ \cite{zhang2012identifiability} and uniformly randomly pick $F$ and $G$ from $(\cdot)^2$, $(\cdot)^3$, and $tanh(\cdot)$ for each repetition.  This is to ensure that our experiment covers a wide range of nonlinear cases. Similarly, the same approach is applied for the relationship between $Y$ and $Z$ with the corresponding function $g(\cdot)$.

And $t=1,2$, $\overline{X}=X_1+X_2$, $\overline{Y}=Y_1+Y_2$, $\overline{Z}=Z_1+Z_2$. And we generate 1000 i.i.d. data points for $(X_1, X_2, Y_1, Y_2, \overline{X}, \overline{Y})$ and feed them into the approximate kernel-based conditional independence test \cite{strobl2019approximate}.  We test the null hypothesis(conditional independence) for (I) $\overline{X} \perp\!\!\!\perp \overline{Y}$, (II) $\overline{Y} \perp\!\!\!\perp \overline{Z}$, (III) $\overline{X} \perp\!\!\!\perp \overline{Z}$, (IV) $\overline{X} \perp\!\!\!\perp \overline{Y} \mid \overline{Z}$, (V) $\overline{Y} \perp\!\!\!\perp \overline{Z} \mid \overline{X}$, (VI) $\overline{X} \perp\!\!\!\perp \overline{Z} \mid \overline{Y}$. And we also test for the conditional independence in corollary \ref{Sufficient Conditions for Conditional Independence}: (A) $\overline{X} \perp\!\!\!\perp Y_1 \mid \overline{Y}$, and (B) $\overline{Z} \perp\!\!\!\perp Y_1 \mid \overline{Y}$. We report the rejection rate, rounded to the nearest percent, at a 5\% significance level in 1000 repeated experiments in Table \ref{experiment result1}, \ref{experiment result2}, \ref{experiment result3}.

\begin{table}[H]
\centering
\caption{Rejection rates for CI tests with different combinations of linear and nonlinear relationships. The index in the box represents the conditional independence that the structure should ideally have. Simply speaking, for the column with the index in the box, the closer the rejection rate is to 5\%, the better. For all other columns from I to VI, a higher rate is better.}
\subfloat[Chain Structure]{
\begin{tabular}{@{}llrrrrrrrr@{}}
\toprule
{$X_{t} \to Y_{t}$} & {$Y_{t} \to Z_{t}$} & {I} & {II} & {III} & {IV} & {V} & {\fbox{VI}} &{A} & {B} \\
\midrule
Linear & Linear & 100\% & 100\% & 100\% & 100\% & 100\% & 6\% & 4\% & 7\%\\
Nonlinear & Linear & 92\% & 100\% & 89\% & 63\% & 100\% & 9\% & 27\% & 10\%\\\
Linear & Nonlinear & 100\% & 95\% & 87\% & 100\% & 94\% & 5\% & 6\% & 82\% \\
Nonlinear & Nonlinear & 92\% & 86\% & 56\% & 89\% & 85\% & 18\% & 27\% & 39\%\\
\bottomrule
\end{tabular}
\label{experiment result1}
}\\
\subfloat[Fork Structure]{
\begin{tabular}{@{}llrrrrrrrr@{}}
\toprule
{$X_{t} \to Y_{t}$} & {$Y_{t} \to Z_{t}$} & {I} & {II} & {III} & {IV} & {V} & {\fbox{VI}} &{A} & {B} \\
\midrule
Linear & Linear & 100\% & 100\% & 100\% & 100\% & 100\% & 5\% & 6\% & 5\%\\
Nonlinear & Linear & 92\% & 100\% & 84\% & 92\% & 100\% & 5\% & 76\% & 5\%\\\
Linear & Nonlinear & 100\% & 93\% & 85\% & 100\% & 93\% & 5\% & 5\% & 71\% \\
Nonlinear & Nonlinear & 92\% & 93\% & 72\% & 86\% & 87\% & 58\% & 72\% & 74\%\\
\bottomrule
\end{tabular}
\label{experiment result2}
}\\
\subfloat[Collider Structure]{
\begin{tabular}{@{}llrrrrrrrr@{}}
\toprule
{$X_{t} \to Y_{t}$} & {$Y_{t} \to Z_{t}$} & {I} & {II} & {\fbox{III}} & {IV} & {V} & {VI} &{A} & {B} \\
\midrule
Linear & Linear & 100\% & 100\% & 5\% & 100\% & 100\% & 99\% & 4\% & 5\% \\
Nonlinear & Linear & 95\% & 89\% & 5\% & 96\% & 91\% & 51\% & 17\% & 56\%\\
Linear & Nonlinear & 90\% & 95\% & 5\% & 91\% & 96\% & 48\% & 54\% & 15\%\\
Nonlinear & Nonlinear & 81\% & 81\% & 6\% & 83\% & 81\% & 29\% & 26\% & 26\%\\
\bottomrule
\end{tabular}
\label{experiment result3}
}
\end{table}

This experiment support our theoretical results, suggesting that conditional independence consistency can be ensured even with some nonlinearity in the model.

Specifically, let's examine the results for chain and fork. We anticipate the tested conditional independence set to contain only $S_{X} \perp\!\!\!\perp S_{Z} \mid S_{Y}$. If so, we can assert that temporal aggregation maintains conditional independence consistency.

From the first and second tables for chain and fork, it's evident that when the model is entirely nonlinear, the results for conditional independence can be erroneous. For instance, the rejection rate for conditional independence that should have been rejected is not high. In the chain structure, the rejection rate for the conditional independence III is zero, implying that every conditional independence test wrongly accepted this conditional independence (type II error). Conversely, the conditional independence that should have been accepted, VI $S_{X} \perp\!\!\!\perp S_{Z} \mid S_{Y}$, has rejection rates of 62\% (chain) and 36\% (fork), significantly exceeding the significance level of 5\%. This aligns with our conclusion in remark \ref{remark: chainlike not consistent in general}, stating that chain and fork models cannot guarantee conditional independence consistency in general cases.

However, when half the model is linear, all conditional independence that should be rejected exhibit high rejection rates, indicating fewer type II errors. Moreover, the rejection rate for the acceptable conditional independence VI is quite low, closely approximating the significance level of 5\%. This suggests that conditional independence-based causal discovery methods can still be applied to temporally aggregated data when the system is partially linear.

Columns A and B primarily aim to validate corollary \ref{corollary:partial linear is enough} and corollary \ref{Sufficient Conditions for Conditional Independence}. The conditional independence represented by A and B corresponds to the two sufficient conditions for conditional independence consistency in corollary \ref{Sufficient Conditions for Conditional Independence}. From the experimental results, we find that if one of these conditions holds, we can ensure conditional independence consistency. For example, in the fork results, under the nonlinear+linear case, B holds while A does not. Nonetheless, we still have conditional independence consistency. Moreover, our findings further verify corollary \ref{corollary:partial linear is enough}, indicating that when a certain part of the causal mechanism is linear, the corresponding sufficient condition in this part is satisfied, ultimately ensuring the conditional independence consistency of the entire system.

Finally, looking at the collider results, conditional independence consistency is maintained under all nonlinear and linear combinations, which agrees with our conclusion in remark \ref{remark: chainlike not consistent in general}, stating that collider can ensure conditional independence consistency under general conditions.

\section{Effect of $k$ Value}\label{Effect of k Value appendix}
\citet{gong2017causal} have proven that the time-delay causal model (time series data) will transform into an instantaneous causal model (i.i.d. data). In our paper, we utilize large $k$ values to approximate the aggregation of the time-delay model as the aggregation of the instantaneous model (aligned model as defined in the main text). We aim to demonstrate the reasonableness of this approximation and to show how quickly the time-delay model transitions to an instantaneous model.

We apply the GES method on a linear fork-like time-delay model (VAR) and a fork-like instantaneous model (aligned model) across different values of $k$. 

\begin{figure}[H]
\centering
\includegraphics[width=0.8\textwidth]{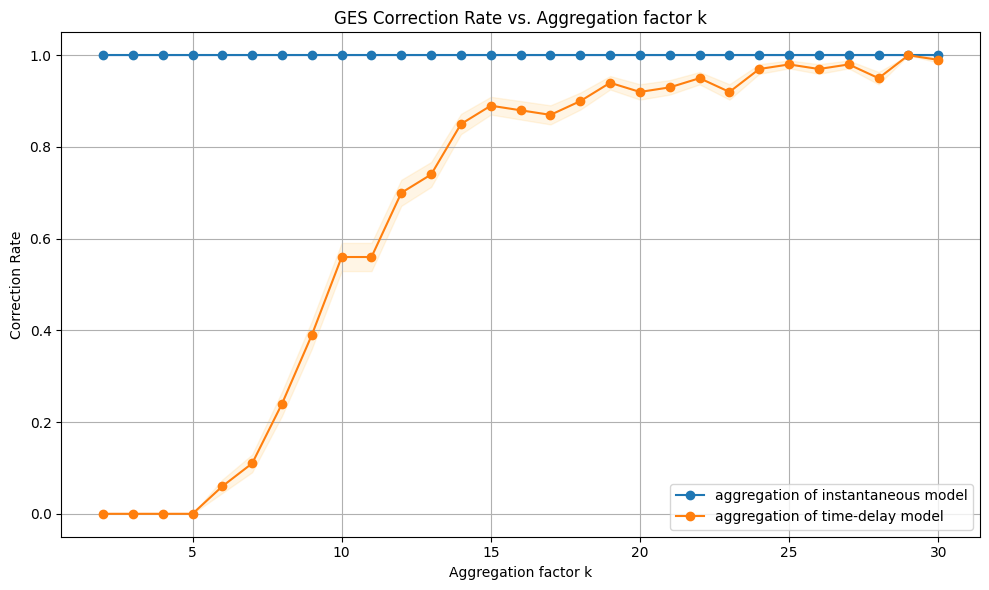}
\caption{GES Accuracy vs. Factor \(k\)}
\label{fig:GES correct rate vs. factor k}
\end{figure}

From Figure \ref{fig:GES correct rate vs. factor k}, we observe that \(k\) does not need to be infinite. A sufficiently large \(k\), such as 20, ensures that the time-delay model becomes an instantaneous model detectable by the instantaneous causal discovery method.

\section{Causal Discovery on Aggregated Data with Prior Knowledge}\label{prior knowledge appendix}
Inspired by Remark \ref{remark: chainlike not consistent in general}, we propose a straightforward solution to ensure that the PC algorithm identifies the correct Markov equivalence. The remark indicates that the collider structure retains conditional independence consistency. This implies that the PC algorithm can determine the correct v-structure if it has the correct skeleton. We therefore conducted experiments to assess whether the PC with a given skeleton as prior knowledge can discern the correct Markov equivalence.

\begin{figure}[H]
\centering
\includegraphics[width=0.8\textwidth]{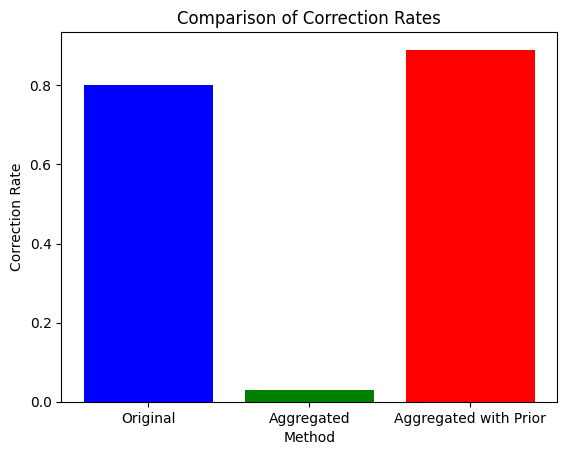}
\caption{Accuracy: PC on Original Data vs. PC on Aggregated Data vs. PC with Prior on Aggregated Data}
\label{fig:Correct Rate: PC on original data vs.}
\end{figure}

\begin{figure}[H]
\centering
\includegraphics[width=0.8\textwidth]{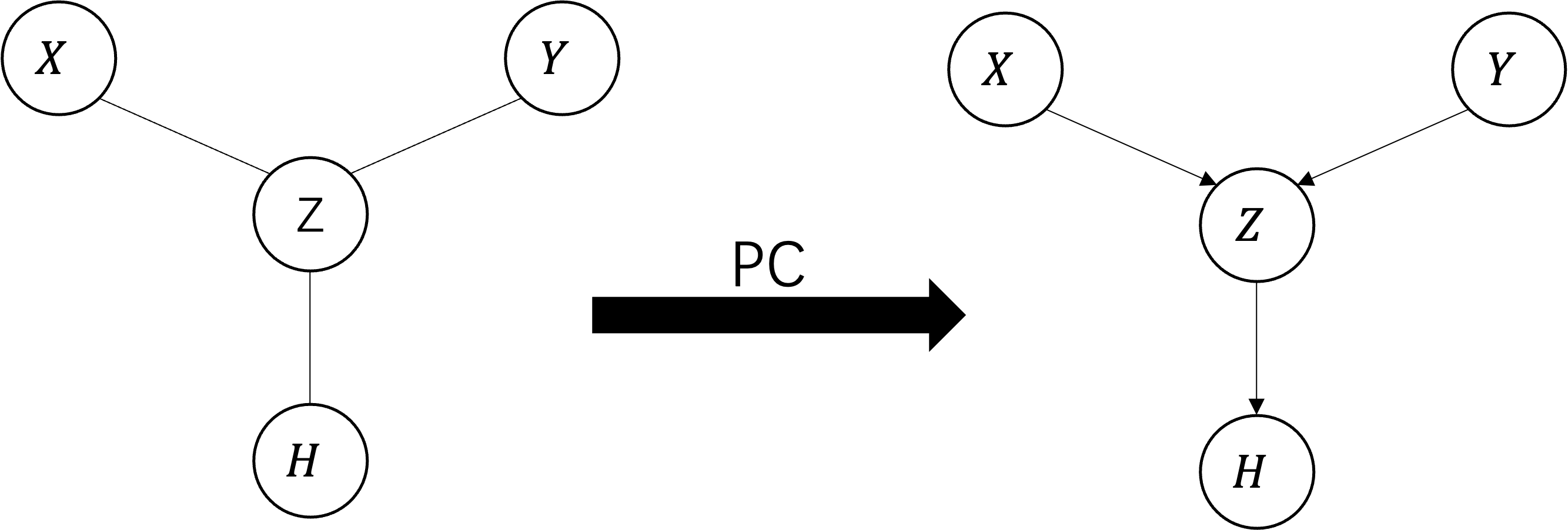}
\caption{PC can find correct v-structure on aggregated data}
\label{fig:pc_with_prior}
\end{figure}

From Figure \ref{fig:Correct Rate: PC on original data vs.}, it becomes evident that when the PC is applied directly to aggregated data, its performance is subpar. However, when provided with the skeleton as prior knowledge, the PC consistently identifies the correct v-structure, yielding accurate results(see Figure \ref{fig:pc_with_prior}). This discovery suggests that future work should concentrate on addressing the aggregation problem during the skeleton discovery process.

\section{Example for Inconsistency in the Nonlinear Additive Noise Model}\label{Example for inconsistency}

Consider the nonlinear additive noise models $X_{t,A}=N_{X,t,A}$, $Y_{t,A} = X_{t,A}^3+N_{t,A}$ and $X_{t,B}=N_{X,t,B}$, $Y_{t,B} =X_{t,B}^3+ N_{t,B}$ for regions A and B, where $t=1,2$, $N_{X,t,A}$ $\sim$ $N(0,1)$, $N_{X,t,B}$ $\sim$ $N(0,2)$, g(k)=1.

According to Theorem 3.3, if we require

$$\overline{Y}_A=\hat{f}(\overline{X}_A)+N_A$$, then it must be

$$\hat{f}(T) = \mathbb{E}\left(\sum_{i=1}^k f(X_i) \mid \overline{X}=T\right) + c$$ here we incorporate c into the noise term, and substituting the definition of the causal function in this example, then we have:

$$\hat{f}(T) = \mathbb{E}\left( X_1^3+X_2^3 \mid X_1+X_2=T\right) $$

And according to the formula for conditional density, we can derive when $X_1$,$X_2$ i.i.d. follow $N(0,\sigma^2)$, and conditioning on $X_1+X_2=T$, the conditional distribution of $X_1$ or $X_2$ is $N(\frac{T}{2},\frac{\sigma^2}{2})$.

And the 3rd raw moment for $N(\mu',\sigma'^2)$ is $\mu'^3+3\mu'\sigma'^2$. Thus, we substitute the conditional expectation and conditional variance to derive (excluding the constant term):

$$\hat{f}(T)=\frac{3T\sigma^2}{4}$$

Therefore,

$$\hat{f}(T)=\frac{3T}{4}$$

for region A, and

$$\hat{f}(T)=3T$$ for region B, which are inconsistent.

\section{Theorem and Proof for Approximating Time-delay Causal Effect by Instantaneous Causal Effect}\label{Theorem and Proof for Approximating}

\begin{theorem}[Asymptotic Equivalence of Time-Delayed and Instantaneous Models]
Given an $s$-dimensional independent noise random vector sequence ${N_0, N_1, N_2, \ldots}$ that are i.i.d. and follow a distribution with zero mean and finite variance, and given that matrix $B$ is an $s\times s$ square matrix with $\| B \| < 1$ (all norms here are $l^2$ norms), and $g(k)$ is always positive and satisfies $\lim_{k \to \infty} g(k) = \infty$.

Consider the time-delayed model:

$$X_0 = N_0$$ $$X_t = BX_{t-1} + N_t$$

and the corresponding instantaneous model:

$$Y_t = BY_t + N_t$$

Then, when $k \to \infty$,

$$\frac{1}{g(k)} \sum_{t=1}^k Y_t - \frac{1}{g(k)} \sum_{t=1}^k X_t \xrightarrow{P} 0$$

\end{theorem}
\begin{proof}
Since $\| B \| < 1$, then $I - B$ is invertible.

Thus,

$$Y_t = (I - B)^{-1} N_t$$

and

 \begin{align} \sum_{t=1}^k Y_t &= \sum_{t=1}^k (I - B)^{-1} N_t \\ &= \sum_{t=1}^k (I + B + B^2 + B^3 + \ldots) N_t \end{align} 

According to

$$X_1 = BX_0 + N_1$$ $$X_2 = BX_1 + N_2 = B^2 N_0 + BN_1 + N_2$$ $$\ldots$$

we have

$$X_t = \sum_{i=0}^t B^{t-i} N_i$$

then

 \begin{align} \sum_{t=1}^k X_t &= \sum_{t=1}^k \sum_{i=0}^t B^{t-i} N_i \\ &= \sum_{i=1}^{k} \sum_{t=i}^{k} B^{t-i} N_i +\sum_{t=1}^k B^{t}N_0\\ &= \sum_{i=1}^{k} \left(\sum_{h=0}^{k-i} B^{h}\right) N_i +\sum_{t=1}^k B^{t}N_0 \text{ (let $h = t-i$)} \end{align} 

Therefore, combining these equations,

$$\frac{1}{g(k)} \sum_{t=1}^k Y_t - \frac{1}{g(k)} \sum_{t=1}^k X_t = \frac{1}{g(k)} \left( \sum_{t=1}^k \left( \sum_{i=k-t+1}^{\infty} B^i \right) N_t - \left( \sum_{i=1}^{k} B^i \right) N_0 \right)$$

$$ \begin{aligned} &\left\| \frac{1}{g(k)} \sum_{t=1}^k Y_t - \frac{1}{g(k)} \sum_{t=1}^k X_t \right\| \\ &\leq \frac{1}{g(k)} \left( \sum_{t=1}^k \left( \sum_{i=k-t+1}^{\infty} \| B \|^i \right) \| N_t \| + \left( \sum_{i=1}^{k} \| B \|^i \right) \| N_0 \| \right) \\ &= \frac{1}{g(k)} \left( \sum_{t=1}^k \frac{\| B \|^{k-t+1}}{1 - \| B \|} \| N_t \| + \frac{\| B \| - \| B \|^{k+1}}{1 - \| B \|} \| N_0 \| \right) \text{ (the sum of a geometric series)} \\ &= \frac{1}{(1 - \| B \|) g(k)} \left( \sum_{t=1}^k \| B \|^{k-t+1} \| N_t \| + (\| B \| - \| B \|^{k+1}) \| N_0 \| \right) \end{aligned} $$

Because ${N_0, N_1, N_2, \ldots}$ are i.i.d. and follow a distribution with zero mean and finite variance, the mean and variance for $\| N_t \|$ are also finite, denoted as $\mu_{\text{norm}}$ and $\sigma_{\text{norm}}^2$, respectively.

Let's prove the expectation and variance of $\frac{1}{(1 - \| B \|) g(k)} \left( \sum_{t=1}^k \| B \|^{k-t+1} \| N_t \| + (\| B \| - \| B \|^{k+1}) \| N_0 \| \right)$ tend to 0 when $k \to \infty$.

Then,

$$ \begin{aligned} &\mathbb{E} \left( \frac{1}{(1 - \| B \|) g(k)} \left( \sum_{t=1}^k \| B \|^{k-t+1} \| N_t \| + (\| B \| - \| B \|^{k+1}) \| N_0 \| \right) \right) \\ &= \frac{1}{(1 - \| B \|) g(k)} \left( \sum_{t=1}^k \| B \|^{k-t+1} \mu_{\text{norm}} + (\| B \| - \| B \|^{k+1}) \mu_{\text{norm}} \right) \\ &= \frac{1}{(1 - \| B \|) g(k)} \left( \frac{\| B \| - \| B \|^{k+1}}{1 - \| B \|} \mu_{\text{norm}} + (\| B \| - \| B \|^{k+1}) \mu_{\text{norm}} \right) \end{aligned} $$

Because $\| B \| < 1$ and $g(k) \to \infty$ as $k \to \infty$, we have

$$ \frac{1}{(1 - \| B \|) g(k)} \left( \frac{\| B \| - \| B \|^{k+1}}{1 - \| B \|} \mu_{\text{norm}} + (\| B \| - \| B \|^{k+1}) \mu_{\text{norm}} \right) \to 0 $$

and

$$ \begin{aligned} &\text{VAR} \left( \frac{1}{(1 - \| B \|) g(k)} \left( \sum_{t=1}^k \| B \|^{k-t+1} \| N_t \| + (\| B \| - \| B \|^{k+1}) \| N_0 \| \right) \right) \\ &= \frac{1}{((1 - \| B \|) g(k))^2} \left( \sum_{t=1}^k \| B \|^{2(k-t+1)} \sigma_{\text{norm}}^2 + (\| B \| - \| B \|^{k+1})^2 \sigma_{\text{norm}}^2 \right) \\ &= \frac{1}{((1 - \| B \|) g(k))^2} \left( \frac{\| B \|^2 - \| B \|^{2k+2}}{1 - \| B \|^2} \sigma_{\text{norm}}^2 + (\| B \| - \| B \|^{k+1})^2 \sigma_{\text{norm}}^2 \right) \end{aligned} $$

Because $\| B \| < 1$ and $g(k) \to \infty$ as $k \to \infty$, we can also conclude the variance tends to 0.

Therefore, according to Chebyshev's inequality, for any $\epsilon > 0$, when $k\to \infty$,

$$P \left( \frac{1}{(1 - \| B \|) g(k)} \left( \sum_{t=1}^k \| B \|^{k-t+1} \| N_t \| + (\| B \| - \| B \|^{k+1}) \| N_0 \| \right) > \epsilon \right) \to 0$$

therefore, for any $\epsilon > 0$, when $k\to \infty$,

$$ \begin{aligned} &P \left( \left\| \frac{1}{g(k)} \sum_{t=1}^k Y_t - \frac{1}{g(k)} \sum_{t=1}^k X_t \right\| > \epsilon \right) \\ &\le P \left( \frac{1}{(1 - \| B \|) g(k)} \left( \sum_{t=1}^k \| B \|^{k-t+1} \| N_t \| + (\| B \| - \| B \|^{k+1}) \| N_0 \| \right) > \epsilon \right) \to 0 \end{aligned} $$

Thus, we conclude convergence in probability.

\end{proof}

\end{document}